\documentclass[lettersize,journal]{IEEEtran}
\usepackage{algorithm,algorithmic}
\usepackage{array}
\usepackage[caption=false,font=footnotesize,labelfont=rm,textfont=rm]{subfig}
\usepackage{textcomp}
\usepackage{stfloats}
\usepackage{url}
\usepackage{verbatim}

\usepackage{graphicx}
\usepackage{cite}
\usepackage{booktabs}
\usepackage{multirow}
\usepackage{epstopdf}
\usepackage{hyperref}
\usepackage{graphicx}
\usepackage{amsmath, amsthm, amsfonts}

\usepackage{soul}
\usepackage{color, xcolor}
\usepackage{siunitx} 
\begin{document}

\title{A Universal Multi-Vehicle Cooperative Decision-Making Approach in Structured Roads\\ by Mixed-Integer Potential Game}

\author{
   Chengzhen Meng, Zhenmin Huang,
	and Jun Ma
    \thanks{Chengzhen Meng, Zhenmin Huang, and Jun Ma are with the Robotics and Autonomous Systems Thrust, The Hong Kong University of Science and Technology, Guangzhou, China, and also the Department of Electronic and Computer Engineering, The Hong Kong University of Science and Technology, Hong Kong SAR, China (e-mail: cmeng403@connect.hkust-gz.edu.cn; zhuangdf@connect.ust.hk; jun.ma@ust.hk). }
	}



\maketitle

\begin{abstract}
Due to the intricate of real-world road topologies and the inherent complexity of autonomous vehicles, cooperative decision-making for multiple connected autonomous vehicles (CAVs) remains a significant challenge. Currently, most methods are tailored to specific scenarios, and the efficiency of existing optimization and learning methods applicable to diverse scenarios is hindered by the complexity of modeling and data dependency, which limit their real-world applicability. To address these issues, this paper proposes a universal multi-vehicle cooperative decision-making method in structured roads with game theory. We transform the decision-making problem into a graph path searching problem within a way-point graph framework. The problem is formulated as a mixed-integer linear programming problem (MILP) first and transformed into a mixed-integer potential game (MIPG), which reduces the scope of problem and ensures that no player needs to sacrifice for the overall cost.
Two Gauss-Seidel algorithms for cooperative decision-making are presented to solve the MIPG problem and obtain the Nash equilibrium solutions. Specifically, the sequential Gauss-Seidel algorithm for cooperative decision-making considers the varying degrees of CAV interactions and flexibility in adjustment strategies to determine optimization priorities, which reduces the frequency of ineffective optimizations. Experimental evaluations across various urban traffic scenarios with different topological structures demonstrate the effectiveness and efficiency of the proposed method compared with MILP and comparisons of different optimization sequences validate the efficiency of the sequential Gauss-Seidel algorithm for cooperative decision-making.
\end{abstract}


\section{Introduction}
\IEEEPARstart{I}{n} recent years, the rapid development of autonomous driving technology has marked a series of significant achievements in this field \cite{omeiza2021explanations,broggi2013extensive,ma2022alternating}, particularly with the emergence of connected autonomous vehicles (CAVs), which are widely regarded as a milestone in the realm of autonomous driving \cite{ZHONG2020102611}. Through vehicle-to-everything (V2X) technology \cite{sun2021survey,noor20226g,abboud2016interworking}, CAVs can communicate with road infrastructure and other vehicles, making the realization of collective intelligence and cooperative decision-making possible. However, cooperative CAV decision-making still faces many challenges. These challenges stem from the inherent non-linearity, non-convexity, and discrete characteristics of cooperative decision-making problems, as well as the varying topological structures associated with different traffic scenarios. To address these difficulties, a considerable amount of work has been proposed \cite{huang2023decentralized,liu2024improved,zhang2021semi}, primarily divided into two categories: learning-based methods and optimization-based methods.

Learning-based methods are known for their wide applicability. For instance, in \cite{li2024flexible}, a cooperative multi-agent reinforcement learning (MARL) algorithm, based on value function decomposition for centralized training and decentralized execution, is proposed. This enables CAVs to quickly pass through mixed traffic flows in emergency situations. In \cite{lv2022cooperative}, a method is introduced for cooperative decision-making in emergencies, based on a deep reinforcement learning (DRL) model to evaluate potential emergency destinations. It also proposes a new representation called the safety evaluation map (SEM) to characterize the evaluation results. In \cite{zhang2023coordinating}, \cite{9507057}, and \cite{meng2017analysis}, learning-based methods are utilized, such as deep learning and multi-agent reinforcement learning, to enhance the operational efficiency and safety of CAVs at intersections. Learning-based approaches can effectively handle complex interactions between CAVs, which are often difficult to model using traditional methods. However, these methods suffer from poor interpretability, which is particularly problematic in the safety-critical field of autonomous driving. Moreover, they rely heavily on specific data for training across different traffic scenarios. Optimization-based methods, unlike learning-based approaches, offer strong interpretability, which are able to find optimal solutions and ensure solution quality. In \cite{duan2023cooperative}, an optimal control model is proposed, which utilizes approximately geometric contour models, dynamic externally tangent rectangle, and inflated rectangle to describe the platoon’s profile. The optimal control model also constructs collision avoidance constraints to ensure safety. However, most methods model the problem as mixed-integer linear programming problem (MILP) \cite{fayazi2018mixed,tajalli2021traffic}, mixed-integer potential game (MIQP) \cite{dollar2021multilane,8736507}, and other NP-hard problems \cite{9222573}, leading to a significant decrease in solving efficiency as the problem size increases. 

In recent years, game theory has been considered an ideal tool for simulating and handling non-cooperative behavior due to its convenience in modeling the interactions between CAVs and considering the selfishness of road users. Unlike optimization-based methods, game theory aims to achieve results that are more satisfactory to all players \cite{qin2024game}. A cooperative decision-making framework based on coalitional game theory is proposed by \cite{hang2021cooperative} for addressing the multi-lane merging problem of CAVs in merging areas, enhancing safety, comfort, and efficiency. In \cite{hang2022decision}, a game-theoretic decision-making framework is designed for CAVs at unsignalized intersection, taking into account both societal and individual interests. In \cite{hang2021decision}, a game-theoretic decision-making framework is introduced to solve the merging, passing, and exiting problems of CAVs in unsignalized roundabout areas, considering both individual benefits of single CAV and the social benefits of the entire traffic system. Game theory and optimization methods have strong interpretability but are often modeled for specific scenarios, addressing issues on a case-by-case basis, such as multi-lane\cite{ LUO201687,amini2021optimizing}, intersection\cite{huang2024non,pan2022convex,rahmati2021helping}, roundabout\cite{medina2022speed,9334596,wei2021autonomous}, etc., which results in weak generalizability.  

To address the aforementioned issues in decision-making of CAVs, this paper proposes an optimization method combined with game theory. It abstracts structured roads and transforms them into directed acyclic graphs (DAGs) to convert the cooperative decision-making problem into a path-finding problem. This is then approximately modeled as an MIPG problem using game theory, and two algorithms are proposed to solve this problem efficiently while ensuring safety and efficiency in driving. 
The main contributions of this paper are fourfold: 

(1) The CAVs cooperative decision-making problem is transformed into a graph path searching problem within a way-point graph framework. We first model the graph path searching problem as an MILP and transform it into an MIPG problem with game theory. The transformation reduces the problem from solving for all CAVs at once to solving for each CAV individually in succession, which maintains a rather small scope of the problem while ensuring that no player needs to sacrifice for the overall cost. 

(2) A Gauss-Seidel algorithm for cooperative decision-making is proposed for solving the MIPG problem, which optimizes the decision variables of each CAV separately during the iteration process rather than considering all of them collectively. Therefore, the Gauss-Seidel algorithm for cooperative decision-making leads to a high solving efficiency compared to MILP methods.

(3) On the basis of the Gauss-Seidel algorithm for cooperative decision-making, a sequential Gauss-Seidel algorithm is further proposed that considers multiple initial factors to determine optimization priorities, and this significantly improves the efficiency of algorithmic solutions. 

(4) The feasibility of the proposed algorithms is verified in three different common structured urban traffic scenarios. The results showcase the generalization and validate the efficiency of the proposed method.

This paper is structured as follows: Section II introduces the problem modeling process, setting constraints, and describing the problem as an MILP problem. Section III elaborates on modeling the problem as an MIPG problem by combining it with game theory, and presents the conditions for equilibrium solutions. Section IV proposes two algorithms for efficiently solving the equilibrium problem. Section V verifies the effectiveness of the proposed methods in three common structured urban traffic scenarios. Finally, we summarize the conclusion and future work in Section VI.

\section{Problem Formulation as Mixed-Integer Linear Programming}
In this section, we propose a universal cooperative decision-making model for CAVs that can be applied to various urban traffic scenarios with structured roads. By discretizing the road network and transforming it into a DAG, the cooperative decision-making problem is transformed into a graph path searching problem, decoupling the relationship between problem formulation and road structure. With the binary decision-making variables and linear constraints, the problem is formulated as an MILP problem. Lane-selection decisions are reflected in the paths on the graph.
\subsection{Way-Point Graph}
In this section, we introduce the concept of the way-point graph in \cite{huang2024universal}. For urban traffic scenarios with structured roads, we create waypoints by sampling equidistantly along the centerline of the road and constructing paths between each waypoint based on reachability. The resulting way-point graph is essentially a DAG. With the introduction of the way-point graph, we can view the decision-making process of CAVs as moving along paths in the graph from the starting point to the desired destination. This approach transforms the decision-making process of CAVs into a path searching problem in a DAG. Since the construction of the way-point graph is related only to the road structure and not affected by the problem's modeling and optimization process, the way-point graph can be pre-constructed.

\begin{figure}[t]
    \includegraphics[width=9cm]{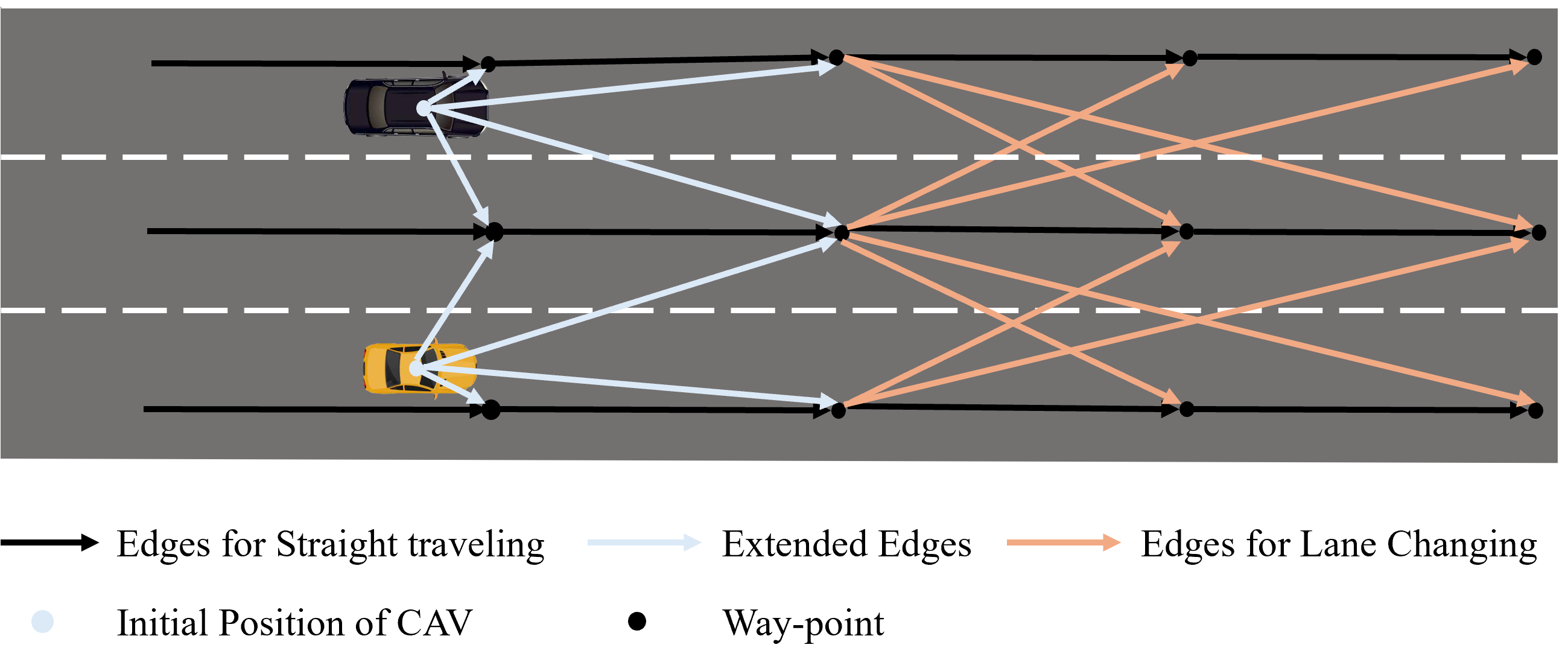}
   \caption{An example of way-point graph. Different arrows represent edges for different driving purposes, including straight traveling, lane changing, etc.  
   }
    \label{Way-point Graph}
\end{figure}

One key issue is that the starting position of the CAV does not completely overlap with the waypoints on the way-point graph, making it impossible to determine the starting point of the CAV on the graph. This problem can be solved by extending the original way-point graph. For each CAV, its starting point is added as a new node to the way-point graph, and this node is connected to the closest fixed number of edges in front of the CAV. An example of this extension is shown in Fig. \ref{Way-point Graph}.

We define the extended way-point graph as $\mathcal{G}=(\mathcal{V} ,\mathcal{E})$, where $\mathcal{V}$ is the set of vertices and $\mathcal{E}$ is the sets of edges. Because of the differences between the starting position of CAVs, which is denoted as the set $\mathcal{N}$, a sub-graph $\mathcal{G}_{i}=(\mathcal{V}_{i},\mathcal{E}_{i})$ is defined for a particular CAV $i$, $i\in\mathcal{N}$. We use ${v}_{1}\longrightarrow {v}_{2}$ to denote if there exists a path from ${v}_{1}\in\mathcal{V}$ to ${v}_{2}\in\mathcal{V}$. Suppose that for CAV $i$, its starting point can be denoted as ${s}_{i}\in\mathcal{V}$, and the set of possible destination can be denoted as ${des}_{i}\subset \mathcal{V}$, then we present the following definitions:
\newtheorem{definition}{Definition}
\begin{definition}
For vertex $i$, we define its forward reachable set $\mathcal{F} \left({v}\right)$ as:
\begin{equation}
    \mathcal{F}\left ({v}\right) =\left \{ {v}_{1} \mid {v}\in \mathcal{V},{v}\longrightarrow {v}_{1} \right \} ,{v} \notin des_{i} 
\end{equation}
and its backward set $\mathcal{B} \left({v}\right)$:
\begin{equation}
    \mathcal{B}\left ({v}\right) =\left \{ {v}_{1}\mid {v}\in \mathcal{V},{v}_{1}\longrightarrow {v} \right \} ,{v}\ne s_{i}. 
\end{equation}
\end{definition}
In most urban road traffic scenarios, reversing is not permitted. Therefore, for a specific CAV $i$, its subgraph should only include a set of reachable vertices that can reach at least one destination vertex, as well as a set of edges composed of these vertices. With the above definitions, the sets of vertices and edges can be expressed as follows:
\begin{equation}
\begin{aligned}
& \mathcal{V}_i=\mathcal{F}\left ( s_i \right ) \cap ( {\textstyle \bigcup_{v\in des_i}}\mathcal{B}\left ( v \right )), \\
& \mathcal{E}_{i}=\left \{ e\mid e\in\mathcal{E},v_{e_{1}}\in \mathcal{V}_{i}, {v}_{e_{2}}\in \mathcal{V}_{i} \right \} 
\end{aligned}
\end{equation}
where $e_{1}$ and $e_{2}$ are the starting point and ending point of edge $e$. An important binary decision variable is introduced as following:
\begin{equation}
P_{i}^{e}= 
\begin{cases}
    1 & \text{if $i$ passes through $e$}, \\
    0 & \text{if $i$ does not pass through $e$}, 
\end{cases}
\quad
\vcenter{\hbox{$\forall i\in \mathcal{N}, \forall e\in \mathcal{E}_i$}}.
\end{equation}

Consequently, all edges traversed by CAV $i$ from the starting point to the destination can be represented by $P_{i}^{e}$. Additionally, to determine the time required for CAV $i$ to pass through all vertices, we define $t_{i}^{v}$ as the time for CAV $i$ to reach vertex $v$. It is important to note that if the CAV does not traverse vertex $v$, the corresponding $t_{i}^{v}$ is considered redundant.

\subsection{Constraints}
\subsubsection{Travel Constraints}
For each CAV $i$, it is essential that it ensures passengers' safety while navigating the road, ensuring both efficiency and comfort. To achieve these objectives, constraints must be imposed on path selection, velocity, longitudinal acceleration, and steering. Additionally, appropriate cost functions need to be designed to penalize inefficient and high-risk behaviors. In this context, we refer to the constraint design outlined in \cite{huang2024universal}, as follows:

\textbf{Path Constraints}: Our goal is to ensure that for each CAV, the paths from the starting point to the destination points are continuous and uninterrupted. Due to the fact that each CAV can only travel on one edge at a time, effective paths must not have branches. To define path constraints, we introduce $\mathcal{E}_{i}^{v,\text{out}}$ and $\mathcal{E}_{i}^{v,\text{in}}$, where $\mathcal{E}_{i}^{v,\text{out}}$ represents the set of outgoing edges from a specific vertex $v$, and $\mathcal{E}_{i}^{v,\text{in}}$ represents the set of incoming edges to a specific vertex $v$. Therefore, to ensure the feasibility of the generated paths, the path constraints are defined as follows:

Starting Point Constraints
\begin{equation}
\label{Path_c_1}
\sum_{e\in \mathcal{E}_{i}^{s_{i},out} }P_{i}^{e}=1,\, \forall i\in\mathcal{N}.
\end{equation}

Destination Constraints
\begin{equation}
\label{Path_c_2}
\sum_{v\in {des_i}}{\sum_{e\in \mathcal{E}_{i}^{v,in}}P_{i}^{e}=1},\, \forall i\in\mathcal{N}.
\end{equation}

Continuity Constraints
\begin{equation}
\label{Path_c_3}
\sum_{e\in \mathcal{E}_{i}^{v,in} } P_{i}^{e}=\sum_{e\in \mathcal{E}_{i}^{v,out}  }P_{i}^{e},\, \forall i\in\mathcal{N},\, \forall v\in\mathcal{\tilde{V}}_{i} 
\end{equation}
where $\mathcal{\tilde{V}}_{i} =\left \{ v\mid v\in\mathcal{V}_{i},v \ne s_i,v\notin des_i \right \} $.

\textbf{Velocity Constraints}:
To meet the requirements for comfort and to avoid safety issues caused by large velocity fluctuations, CAVs should maintain a stable velocity as much as possible. Therefore, we set a reference velocity $V_{i}^{r}$ for each CAV. For the edge $e$ that CAV $i$ passes through, we define the velocity deviation as $\Delta V_{i}^{e}= V_{i}^{e}- V_{i}^{r}$, where $V_{i}^{e}$ is the average velocity of the CAV $i$ on edge $e$. To reflect the velocity deviation of $i$ on edge $e$ in the time domain $\left [ t_{i}^{e_1},t_{i}^{e_2} \right ]$, where $e_{1}$ and $e_{2}$ are the starting point and ending point of edge $e$, we aim to minimize $\left |\triangle V_{i}^{e}\left (  t_{i}^{e_2}-t_{i}^{e_1} \right )  \right | $ and the constraints can be expressed as:
\begin{equation}
\begin{aligned}
\label{V_c_1}
& l^{e}-V_{i}^{r}\left (t_{i}^{e_2}-t_{i}^{e_1} \right )\le M\left (1-P_{i}^{e} \right )+\overline{s}_{i}^{e} \\
& l^{e}-V_{i}^{r}\left (t_{i}^{e_2}-t_{i}^{e_1} \right )\ge -M\left (1-P_{i}^{e} \right )-\underline{s}_{i}^{e} \\
& \forall i\in\mathcal{N},\, \forall e\in\mathcal{E}_{i}
\end{aligned}
\end{equation}
where  $l^{e}$ is the length of edge $e$, and $M$ is a large constant used in the big-M method \cite{bazaraa2011linear}. $[\overline{s}_{i}^{e},\underline{s}_{i}^{e}]$ is a pair of slack variables to penalize the changes of velocity. To ensure that the average velocity $V_{i}^{e}$ is bounded, which means $V_{i}^{min}\le V_{i}^{e}\le V_{i}^{max}$, [$\overline{s}_{i}^{e}$, $\underline{s}_{i}^{e}$] need to satisfy the following constraints:
\begin{equation}
\begin{aligned}
\label{V_c_2}
& 0 \le \overline{s}_{i}^{e} \le \left (V_{i}^{max}-V_{i}^{r} \right )\left (t_{i}^{e_2}-t_{i}^{e_1} \right )+M\left (1-P_{i}^{e} \right ) \\
& 0 \le \underline{s}_{i}^{e} \le \left (V_{i}^{r}-V_{i}^{min} \right )\left (t_{i}^{e_2}-t_{i}^{e_1} \right )+M\left (1-P_{i}^{e} \right ) \\
& \forall i\in\mathcal{N},\, \forall e\in\mathcal{E}_{i}
\end{aligned}
\end{equation}
where $V_{i}^{max}$ and $V_{i}^{min}$ represent the upper and lower bounds of the velocity of CAV $i$.

\textbf{Longitudinal Acceleration Constraints}: To enhance passenger comfort and avoid excessive longitudinal acceleration, it is essential to constrain the acceleration. We consider the acceleration at vertices by representing it through the velocities of the edges before and after each vertex. By linearizing this approximation, we establish the necessary constraints. The detailed derivation process can be found in \cite{huang2024universal}. Here, we present only the final results.

We divide the interval $\left[V_{i}^{\text{min}}, V_{i}^{\text{max}} \right]$ into $k$ regions and set a reference velocity $V_{i,k}^r$ for each region. Let $y_{i,k}^{\alpha\beta}$ be a binary variable which is 1 if the velocity belongs to a certain region $[V_{i,k}^{\text{min}}, V_{i,k}^{\text{max}}]$, and 0 otherwise. The constraints can be expressed as follows:

\begin{equation}
\begin{aligned}
\label{a_c_1}
        & \sum_{k}y_{i,k}^{v,\alpha\beta} =1, \\
        & \sum_{k}y_{i,k}^{v,\alpha\beta} \left ( \frac{l^\alpha +l^\beta}{V_{i,k}^{min}}  \right ) \ge t_i^{\beta_2}-t_i^{\alpha_1}-M\left ( 2-P_i^{\alpha }-P_i^{\beta } \right ), \\
        & \sum_{k}y_{i,k}^{v,\alpha\beta} \left ( \frac{l^\alpha +l^\beta}{V_{i,k}^{max}}  \right ) \le t_i^{\beta_2}-t_i^{\alpha_1}+M\left ( 2-P_i^{\alpha }-P_i^{\beta } \right ), \\
        & \forall i \in \mathcal{N},\forall v \in \check{\mathcal{V}}_i,\alpha \in \mathcal{E}_i^{v,in},\beta \in \mathcal{E}_i^{v,out}.
\end{aligned}
\end{equation}

With the slack variables $\left[\overline{\gamma}_{i,k}^{v,\alpha\beta},\underline{\gamma}_{i,k}^{v,\alpha\beta}\right]$ to penalize the longitudinal acceleration, the constraints are introduced as:
\begin{align}
\begin{split}
\label{a_c_2}
        & \frac{t_i^{\alpha_1} -t_i^v}{l^\alpha } -\frac{t_i^{\beta_2} -t_i^v}{l^\beta } \le M\left ( 3-P_i^{\alpha }-P_i^{\beta }-y_{i,k}^{v,\alpha\beta} \right ) +\overline{\gamma}_{i,k}^{v,\alpha\beta},\\
        & \frac{t_i^{\alpha_1} -t_i^v}{l^\alpha } -\frac{t_i^{\beta_2} -t_i^v}{l^\beta } \le \!-M\!\left ( 3-P_i^{\alpha }-P_i^{\beta }-y_{i,k}^{v,\alpha\beta} \right )\!-\! \underline{\gamma}_{i,k}^{v,\alpha\beta}, \\
        &\overline{\gamma}_{i,k}^{v,\alpha\beta} \le \frac{\gamma_{max}\left ( t_i^{\beta_2}-t_i^{\alpha_1} \right )}{2\left ( V_{i,k}^{r}  \right )^2 } +M\left ( 3-P_i^{\alpha }-P_i^{\beta }-y_{i,k}^{v,\alpha\beta} \right ), \\
        &\underline{\gamma}_{i,k}^{v,\alpha\beta} \le -\frac{\gamma_{min}\left ( t_i^{\beta_2}-t_i^{\alpha_1} \right )}{2\left ( V_{i,k}^{r}   \right )^2 } +M\left ( 3-P_i^{\alpha }-P_i^{\beta }-y_{i,k}^{v,\alpha\beta} \right ), \\
        & \overline{\gamma}_{i,k}^{v,\alpha\beta},   \underline{\gamma}_{i,k}^{v,\alpha\beta} \ge 0, \\
        & \forall i \in \mathcal{N},\forall v \in \check{\mathcal{V}}_i,\alpha \in \mathcal{E}_i^{v,in},\beta \in \mathcal{E}_i^{v,out}
\end{split}
\end{align}
where $\gamma_{min}$ and $\gamma_{max}$ are constants, denoting the minimum and maximum of deceleration and acceleration, respectively. 

For the starting point, the constraints are adjusted accordingly due to the lack of incoming edges. We denotes $V_i^{ini}$ as the initial velocity of CAV $i$, then the constraints can be expressed as:
\begin{equation}
\begin{aligned}
\label{a_c_3}
        & \sum_{k}y_{i,k}^{{s_i},\beta} =1, \\
        & \sum_{k}y_{i,k}^{{s_i},\beta} \left ( \frac{l^\beta}{V_{i,k}^{min}}  \right ) \ge t_i^{\beta_2}-M\left ( 1-P_i^{\beta } \right ), \\
        & \sum_{k}y_{i,k}^{{s_i},\beta} \left ( \frac{l^\beta}{V_{i,k}^{max}}  \right ) \le t_i^{\beta_2}+M\left ( 1-P_i^{\beta } \right ), \\
        & \forall i \in \mathcal{N},\beta \in \mathcal{E}_i^{{s_i},out}.
\end{aligned}
\end{equation}

\begin{equation}
\begin{aligned}
\label{a_c_4}
        & \frac{2V_{i,k}^r-V_i^{ini}}{\left ( V_{i,k}^r \right )^2 } -\frac{t_i^{\beta_2}}{l^\beta} \le M\left ( 2-P_i^\beta -y_{i,k}^{{s_i},\beta} \right )+\overline{\gamma}_{i,k}^{{s_i},\beta}, \\
        & \frac{2V_{i,k}^r-V_i^{ini}}{\left ( V_{i,k}^r \right )^2 } -\frac{t_i^{\beta_2}}{l^\beta} \ge -M\left ( 2-P_i^\beta -y_{i,k}^{{s_i},\beta} \right )-\underline{\gamma}_{i,k}^{{s_i},\beta}, \\
        & 0 \le \overline{\gamma}_{i,k}^{{s_i},\beta} \le \frac{\gamma_{max}t_i^{\beta_2}}{2\left ( V_{i,k}^r\right )^2 }+M\left ( 2-P_i^{\beta }-y_{i,k}^{{s_i},\beta} \right ), \\
         & 0 \le \underline{\gamma}_{i,k}^{{s_i},\beta} \le -\frac{\gamma_{min}t_i^{\beta_2}}{2\left ( V_{i,k}^r  \right )^2 }+M\left ( 2-P_i^{\beta }-y_{i,k}^{{s_i},\beta} \right), \\
        & \forall i \in \mathcal{N},\beta \in \mathcal{E}_i^{v,out}. 
\end{aligned}
\end{equation}

\textbf{Steering Constraints}: To avoid frequent lane changes or high-speed turns, it is imperative to impose constraints on steering. In most cases, CAVs generate lateral acceleration when turning. Therefore, we constrain the lateral acceleration $a_i^{lat}$. To properly reflect the impact of steering on driving, we choose to penalize the integral of lateral acceleration, which corresponds to steering angle and speed. Since steering only occurs at vertices in our proposed method, the integral time span $\left[t_i^{\alpha_1}, t_i^{\beta_2}\right]$ represents the time between the two edges $\alpha$ and $\beta$, which are the incoming and outgoing edges, respectively, passing through a vertex $v$. We define the angle between the two edges $\alpha$ and $\beta$ as $\theta^{\alpha\beta}$. The constraints can be expressed as:

\begin{equation}
    \begin{aligned}
    \label{Ro_c_1}
        & V_{i,k}^{r}\theta^{\alpha\beta}-M\left ( 3-P_i^{\alpha }-P_i^{\beta }-y_{i,k}^{v,\alpha\beta} \right ) \le \eta _{i,k}^{v,\alpha \beta }, \\
        & 0 \le \eta _{i,k}^{v,\alpha \beta} \le  \eta_{max}\left (t_i^{\beta_2} -t_i^{\alpha_1} \right )+M\left ( 3-P_i^{\alpha }-P_i^{\beta }-y_{i,k}^{v,\alpha\beta} \right ),\\
        & \forall i \in \mathcal{N},\forall v \in \check{\mathcal{V}}_i,\alpha \in \mathcal{E}_i^{v,in},\beta \in \mathcal{E}_i^{v,out} 
    \end{aligned}
\end{equation}
where $\eta _{i,k}^{v,\alpha \beta}$ is the slack variable to be penalized in the cost, and $\eta_{max}$ is a pre-defined constant to bound the lateral acceleration and avoid high-speed turns. For the starting point, the constraints can be expressed as:
\begin{equation}
    \begin{aligned}
    \label{Ro_c_2}
        & V_{i,k}^{r}\theta^{init\beta}-M\left ( 2-P_i^{\beta }-y_{i,k}^{s_i,\beta} \right ) \le \eta _{i,k}^{s_i,\beta}, \\
        & 0 \le \eta _{i,k}^{s_i,\beta} \le  \eta_{max}t_i^{\beta_2}+M\left ( 2-P_i^{\beta }-y_{i,k}^{s_i,\beta} \right ), \\
        & \forall i \in \mathcal{N},\beta \in \mathcal{E}_i^{s_i,out} 
    \end{aligned}
\end{equation}
where $\theta^{init\beta}$ is the angle between the initial heading of CAV $i$ and edge $\beta$.

\subsubsection{Time Constraints}
In urban road scenarios, reversing while driving is generally not permissible. Consequently, the time at points closer to the CAV in the direction of travel must be less than the time at points farther away. To standardize the initial time of all CAVs, we set the initial time for any CAV $i$ to be $0$, and ensure that the time at all points is non-negative. These constraints can be expressed as:
\begin{equation}
\label{t_c_1}
\begin{aligned}
& t_i^{s_i}=0, \\ 
& t_i^{e_1} < t_i^{e_2},\forall i \in \mathcal{N}, \forall e \in \mathcal{E},  \\
& t_i^v \ge 0.
\end{aligned}
\end{equation}
\subsubsection{Collision Avoidance Constraints}
For the purpose of preventing any potential collision accidents and ensuring driving safety, constraints are imposed on CAV $i$ and its surrounding CAV $j$. Different from the method in \cite{huang2024universal}, in this section, we design collision avoidance constraints for individual CAV as the basis for optimizing each CAV in the subsequent algorithm. Here, we assume that the trajectories of other CAVs $j$ are known when considering collision constraints for CAV $i$. To avoid collisions, we first introduce the concept of crucial edge pairs $\left \{ \left ( i,e_i \right ),\left ( j,e_j \right ) \right \} $, and crucial areas $C_{ij}^{e_{i}e_{j}}$ as in \cite{huang2024universal}. For CAV $i$ and $e_{i} \in \mathcal{E}_{i}$, we define the set of surrounding CAVs as $\mathcal{N}_{i}$ and consider the situation where the front of $i$ is parallel to $e$. The volume occupied by CAV $i$ when it is aligned with $e_i$ is defined as $S_{i}^{e_{i}}\left ( \theta  \right )$ and its location is at $\left ( 1-\theta  \right )p_{i}^{e_{1}}+\theta p_{i}^{e_{2}}$, where $p_{i}^{e_{1}}$ and $p_{i}^{e_{2}}$ represent the locations of starting point and end point, respectively. Based on this, we introduce the following definitions:

 \begin{definition}
 For $i\in\mathcal{N}$,$j\in\mathcal{N}_i$, $e_{i}\in\mathcal{E}_{i}$, and $e_{j}\in\mathcal{E}_{j}$, $\left \{ \left ( i,e_i \right ),\left ( j,e_j \right ) \right \} $ forms a crucial edge pair if ${S}_{i}^{e}\cap {S}_{j}^{e}\ne 0$. The set of all the crucial edge pairs is defined as $\mathcal{R}$.
 \end{definition}


Based on this, the crucial area $C_{ij}^{e_{i}e_{j}}$ can be defined as:
\begin{equation}
\begin{aligned}
& \theta_{ij,1}^{e_{i}e_{j}}=\min_{0\le \theta\le 1 }\left \{ S_{i}^{e_{i}}\left ( \theta  \right )\cap S_{j}^{e_{j}} \ne 0 \right\}, \\
& \theta_{ij,2}^{e_{i}e_{j}}=\max_{0\le \theta\le 1 }\left \{ S_{i}^{e_{i}}\left ( \theta  \right )\cap S_{j}^{e_{j}} \ne 0 \right \}, \\
& C_{ij}^{e_{i}e_{j}}=\left \{ \left ( 1-\theta  \right )p_{i}^{e_{1}}+\theta p_{i}^{e_{2}} \mid
\theta_{ij,1}^{e_{i}e_{j}}\le \theta \le\theta_{ij,2}^{e_{i}e_{j}} \right\}
\end{aligned}
\end{equation}
where $\theta_{ij,1}^{e_{i}e_{j}}$ and $\theta_{ij,2}^{e_{i}e_{j}}$ represent the boundary position of the crucial area.

Obviously, the crucial area can be determined before the optimization process. Assuming that the motion of CAVs along a road edge can be approximated as uniform motion, the time $t_{i,\theta}^{e_{i}}$ at position $p_{\theta}^{e_{i}}$ can be obtained through linear interpolation:

\begin{equation}
\label{time_inter}
t_{i,\theta}^{e_{i}}\approx \left ( 1-\theta  \right )t_{i}^{e_{i,1}}+\theta t_{i}^{e_{i,2}}.
\end{equation}
For CAV $j$, we define the time it enters the crucial area as $t_{ij,1}^{e_{i}e_{j}}$ and the time it exits as $t_{ij,2}^{e_{i}e_{j}}$. With (\ref{time_inter}), the time CAV $j$ enters the crucial area can be expressed as:
\begin{equation}
\begin{aligned}
& t_{ij,1}^{e_{i}e_{j}}\approx \left ( 1-\theta_{ij,1}^{e_{i}e_{j}}  \right )t_{j}^{e_{j,1}}+\theta_{ij,1}^{e_{i}e_{j}} t_{j}^{e_{j,2}}, \\
& t_{ij,2}^{e_{i}e_{j}}\approx \left ( 1-\theta_{ij,2}^{e_{i}e_{j}}  \right )t_{j}^{e_{j,1}}+\theta_{ij,2}^{e_{i}e_{j}} t_{j}^{e_{j,2}}.  
\end{aligned}
\end{equation}

Due to the different scenarios in which crucial edge pairs and crucial areas can occur, to reflect generality and obtain conditions to avoid collisions in all scenarios, we project the movements of CAV $j \in \mathcal{N}_i$ onto the direction of motion of CAV $i$, thereby reducing the original two-dimensional motion to one dimension. At the same time, the distance between these two CAVs should be greater than the equivalent safe distance.

For the crucial edge pair $\left \{ \left ( i,e_i \right ),\left ( j,e_j \right ) \right \} $, we project $\left ( j,e_j \right )$ onto the corresponding $\left(i, e_i \right)$, where the projected starting and ending points of the crucial area are denoted as $\hat{s}_{ij,1}^{e_i e_j}$ and $\hat{s}_{ij,2}^{e_i e_j}$. $L_i$ and $\hat{L}_j$ are defined as the length of CAV $i$ and the projected length of CAV $j$ on the trajectory of CAV $i$, respectively. The equivalent safe distance can be expressed as $D_{ij} = \frac{1}{2} \left(L_i + \hat{L}_j \right)$. To cover all scenarios, we discuss four different cases:

\begin{itemize}
	\item Case 1.1: The angle between $e_i$ and $e_j$ is less than $90^{\circ}$ and CAV $i$ passes the crucial area in front of CAV $j$.
 	\item Case 1.2: The angle between $e_i$ and $e_j$ is less than $90^{\circ}$ and CAV $i$ passes the crucial area behind CAV $j$.
	\item Case 2.1: The angle between $e_i$ and $e_j$ is greater than or equal to $90^{\circ}$ CAV $i$ passes the crucial area in front of CAV $j$.
 	\item Case 2.2: The angle between $e_i$ and $e_j$ is greater than or equal to $90^{\circ}$ CAV $i$ passes the crucial area behind CAV $j$.
\end{itemize}

 Based on the above definitions and transformations, we derive the condition to avoid collisions: during the process of CAV $j$ passing through $\hat{s}_{ij,1}^{e_i e_j}$ to $\hat{s}_{ij,2}^{e_i e_j}$, CAV $i$ must always be either ahead of or behind CAV $j$. To fully cover all cases for setting the constraints, we introduce a new binary variable $\delta_{ij}^{e_i}$ to describe the relative positional relationship between CAV $i$ and CAV $j$:

 \begin{equation}
\delta_{ij}^{e_i}= 
\begin{cases}
    1 & \text{if $i$ in front of $j$}, \\
    0 & \text{if $i$ behind $j$}, 
\end{cases}
\quad
\vcenter{\hbox{$\forall i\in \mathcal{N}, \forall e_{i}\in \mathcal{R}$}}.
\end{equation}

\begin{figure}[htbp]
\centering
\subfloat[Case 1.1]{\includegraphics[width=0.48\linewidth]{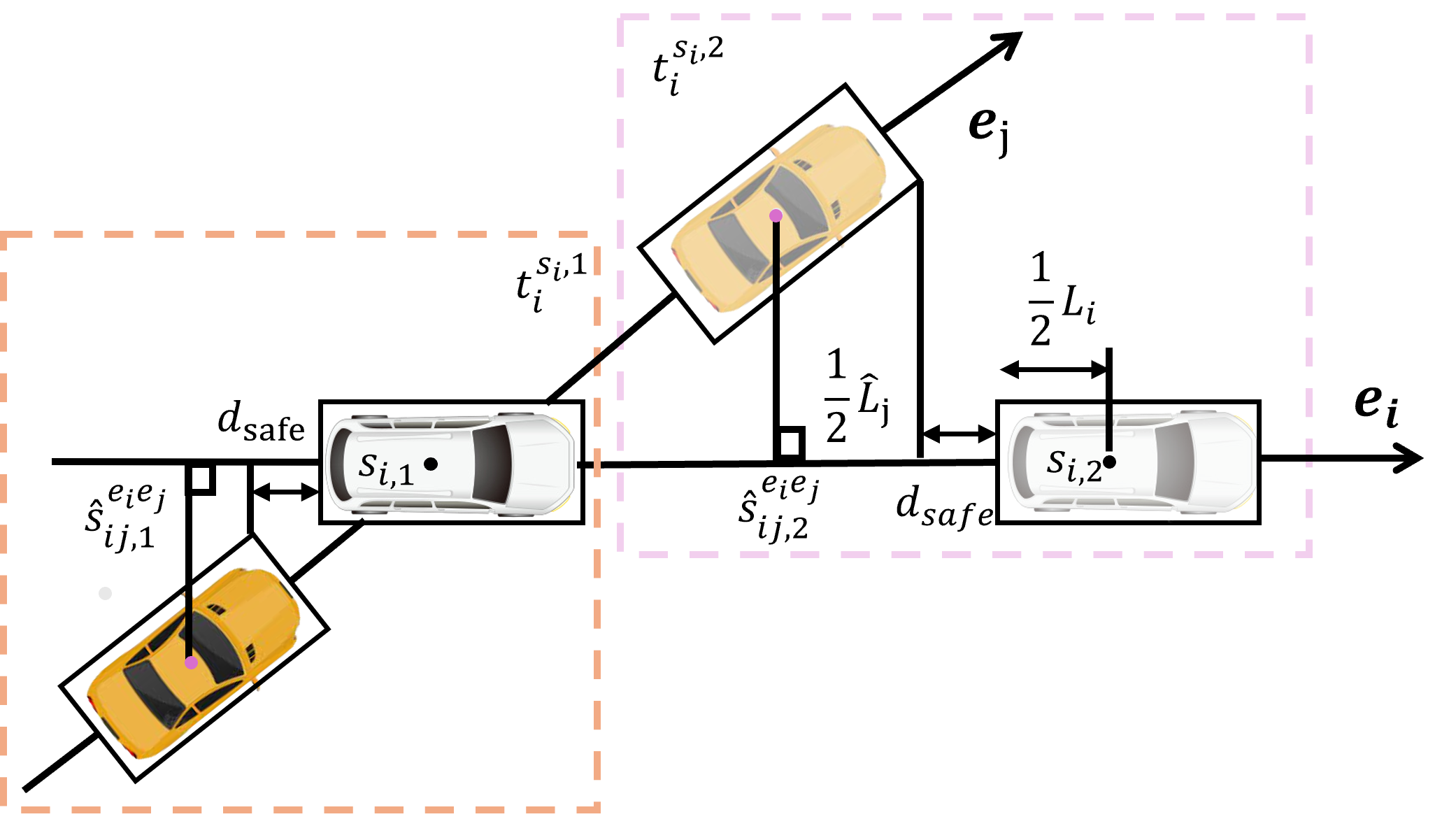}
\label{Case1.1}}
\subfloat[Case 1.2]{\includegraphics[width=0.48\linewidth]{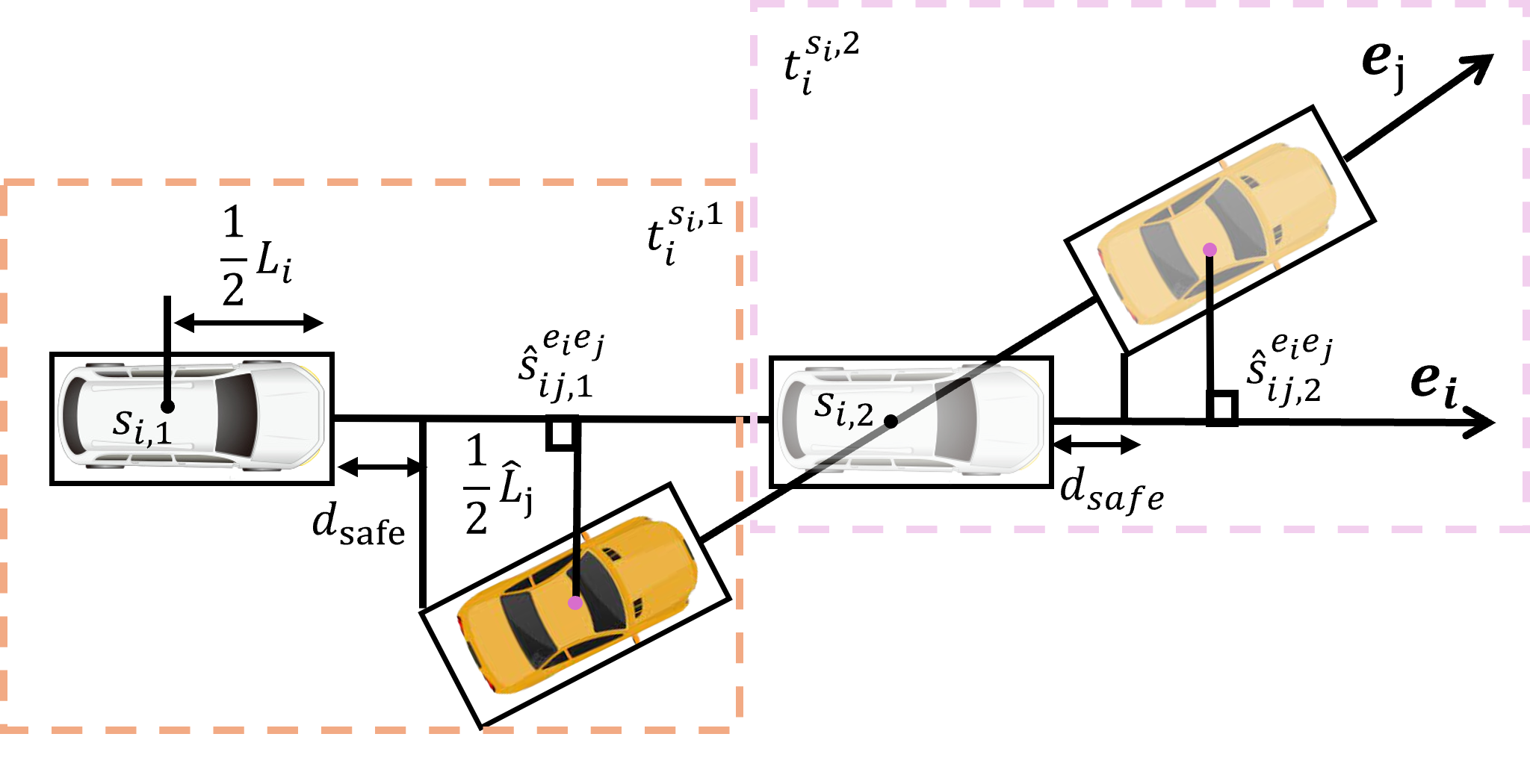}
\label{Case1.2}}
\vspace{5mm}
    \subfloat[Case 2.1]{\includegraphics[width=0.48\linewidth]{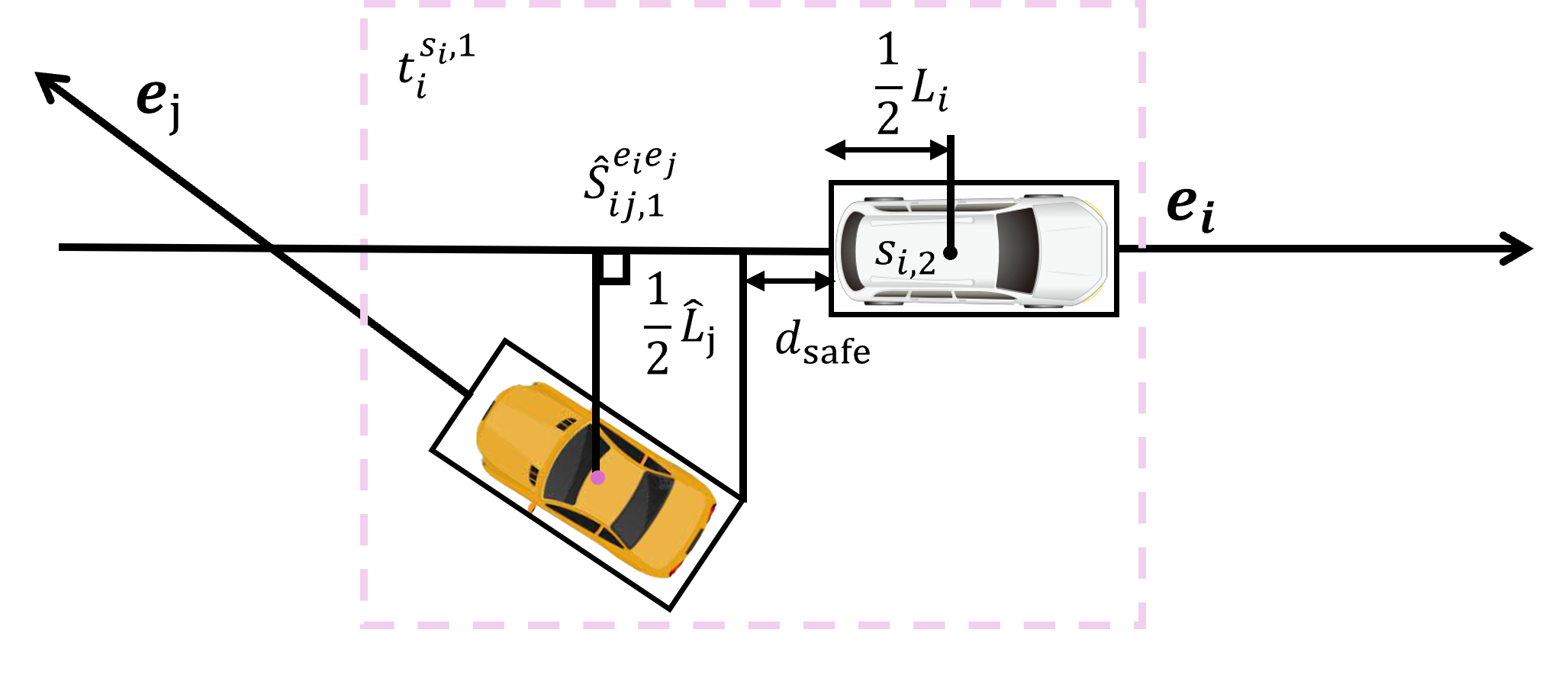}\label{Case2.1}}
    \subfloat[Case 2.2]{\includegraphics[width=0.48\linewidth]{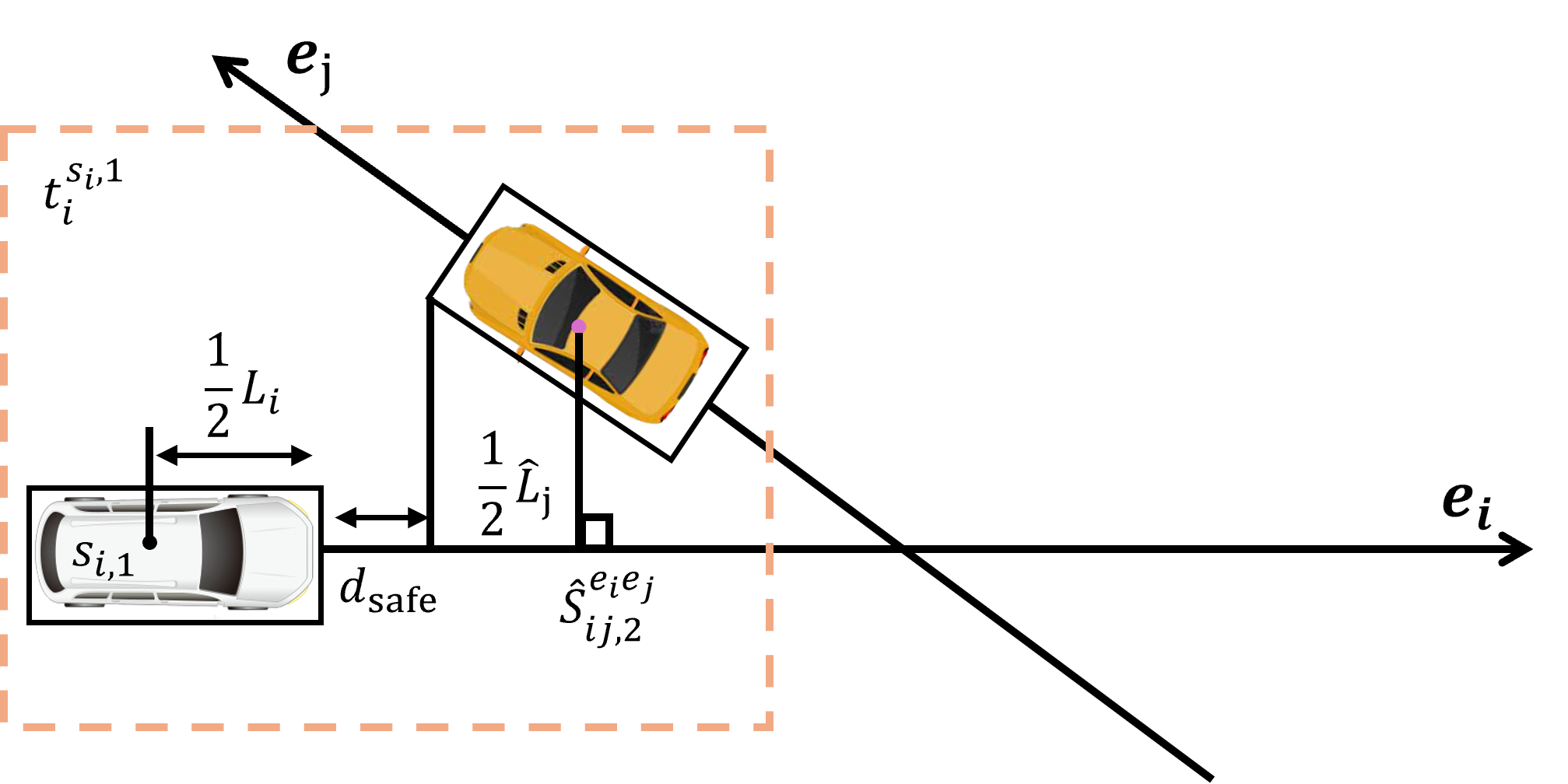}\label{Case2.2}}
    \caption{Projection of each point and the relationship of CAV $i$ and CAV $j$ positions.}
    \label{Case}
\end{figure}

For Case 1.1 and Case 1.2, the positional relationship between CAV $i$ and CAV $j$ is illustrated in Fig. \ref{Case}\subref{Case1.1}-\subref{Case1.2}. Since the movement of CAVs is represented using the positions of their center points, this safe distance is equivalent to the previously mentioned $D_{ij}$.

When CAV $j$ moves to the equivalent edge positions $\hat{s}_{ij,1}^{e_i e_j}$ and $\hat{s}_{ij,2}^{e_i e_j}$ of the crucial area, we define the limit positions for CAV $i$ as $s_{i,1}$ and $s_{i,2}$, as shown in Fig. \ref{Case}, and the corresponding times for these positions as $t_i^{s_{i,1}}$ and $t_i^{s_{i,2}}$. For all cases, the method to calculate $t_i^{s_{i,1}}$ and $t_i^{s_{i,2}}$ is the same and can be expressed as:

\begin{equation}
\begin{aligned}
& t_i^{s_i,1} = \left ( 1-s_{i,1}  \right )t_{i}^{e_{i,1}}+s_{i,1}t_{i}^{e_{i,2}}, \\
& t_i^{s_i,2} = \left ( 1-s_{i,2} \right )t_{i}^{e_{i,1}}+s_{i,2}t_{i}^{e_{i,2}}.
\end{aligned}
\end{equation}
For Case 1.1, the limit positions can be expressed as:
\begin{equation}
\begin{aligned}
& s_{i,1} = \frac{\hat{s}_{ij,1}^{e_{i}e_{j}}+D_{ij}}{l^{e_i}}, \\
& s_{i,2} = \frac{\hat{s}_{ij,2}^{e_{i}e_{j}}+D_{ij}}{l^{e_i}}.  
\end{aligned}
\end{equation}
A sufficient condition for avoiding collision, as shown in Fig. \ref{Case}\subref{Case1.1}, can be expressed as:
\begin{equation}
\begin{aligned}
\label{C_C_1}
& t_i^{s_{i,1}} \le  t_{ij,1}^{e_{i}e_{j}}+M\left ( 1-\delta _{ij}^{e_i} \right ), \\
& t_i^{s_{i,2}} \le  t_{ij,2}^{e_{i}e_{j}}+M\left ( 1-\delta _{ij}^{e_i} \right ). 
\end{aligned}
\end{equation}
For Case 1.2, similarly, it can be expressed as:
\begin{equation}
\begin{aligned}
& s_{i,1} = \frac{\hat{s}_{ij,1}^{e_{i}e_{j}}-D_{ij}}{l^{e_i}}, \\
& s_{i,2} = \frac{\hat{s}_{ij,2}^{e_{i}e_{j}}-D_{ij}}{l^{e_i}}. 
\end{aligned}
\end{equation}
A sufficient condition for avoiding collision, as shown in Fig. \ref{Case}\subref{Case1.2}, can be expressed as:
\begin{equation}
\begin{aligned}
\label{C_C_2}
& t_i^{s_{i,1}}\ge t_{ij,1}^{e_{i}e_{j}}-M\delta _{ij}^{e_i},\\
& t_i^{s_{i,2}}\ge t_{ij,2}^{e_{i}e_{j}}-M\delta _{ij}^{e_i}.  
\end{aligned}
\end{equation}

For Case 2.1 and Case 2.2, the position relationship between CAV $i$ and CAV $j$ is as shown in the Fig. \ref{Case}\subref{Case2.1}-\subref{Case2.2}. For Case 2.1, it is a little different from Case 1.1 and Case 1.2. In this condition, we just need to take one limit position into consideration, which can be expressed as:
\begin{equation}
s_{i,2} = \frac{\hat{s}_{ij,1}^{e_{i}e_{j}}+D_{ij}}{l^{e_i}}.  
\end{equation}
A sufficient condition for avoiding collision in the corresponding condition, as shown in Fig. \ref{Case}\subref{Case2.1}, can be expressed as:
\begin{equation}
\label{C_C_3}
t_i^{s_{i,2}} \le  t_{ij,1}^{e_{i}e_{j}}+M\left ( 1-\delta _{ij}^{e_i} \right ). 
\end{equation}
Similarly, for Case 2.2, as shown in Fig. \ref{Case}\subref{Case2.2}, the constraints can be expressed as:
\begin{equation}
\begin{aligned}
\label{C_C_4}
& s_{i,1} = \frac{\hat{s}_{ij,2}^{e_{i}e_{j}}+D_{ij}}{l^{e_i}},\\
& t_i^{s_{i,1}}\ge t_{ij,2}^{e_{i}e_{j}}-M\delta _{ij}^{e_i} . 
\end{aligned}
\end{equation}

Gathering all the cases we discussed above, the constraints can be expressed as:
\begin{equation}
\begin{aligned}
\label{C_C_T1}
& \left ( 1-\frac{\hat{s}_{ij,1}^{e_{i}e_{j}}-D_{ij}}{l^{e_i}}  \right )t_{i}^{e_{i,1}}+\frac{\hat{s}_{ij,1}^{e_{i}e_{j}}-D_{ij}}{l^{e_i}}t_{i}^{e_{i,2}} \ge \left ( 1-\theta_{ij,1}^{e_{i}e_{j}}  
 \right )t_{j}^{e_{j,1}}\\
&+\theta_{ij,1}^{e_{i}e_{j}} 
 t_{j}^{e_{j,2}}-M\delta _{ij}^{e_i},\\
& \left ( 1-\frac{\hat{s}_{ij,2}^{e_{i}e_{j}}-D_{ij}}{l^{e_i}}  \right )t_{i}^{e_{i,1}}+\frac{\hat{s}_{ij,2}^{e_{i}e_{j}}-D_{ij}}{l^{e_i}}t_{i}^{e_{i,2}} \ge \left ( 1-\theta_{ij,2}^{e_{i}e_{j}}  \right )t_{j}^{e_{j,1}}\\
&+\theta_{ij,2}^{e_{i}e_{j}} t_{j}^{e_{j,2}}-M\delta _{ij}^{e_i},\\
& \left ( 1-\frac{\hat{s}_{ij,1}^{e_{i}e_{j}}+D_{ij}}{l^{e_i}}  \right )t_{i}^{e_{i,1}}+\frac{\hat{s}_{ij,1}^{e_{i}e_{j}}+D_{ij}}{l^{e_i}}t_{i}^{e_{i,2}} \le \left ( 1-\theta_{ij,1}^{e_{i}e_{j}}  \right )t_{j}^{e_{j,1}}\\
&+\theta_{ij,1}^{e_{i}e_{j}} t_{j}^{e_{j,2}}+M\left(1-\delta _{ij}^{e_i}\right),\\
& \left ( 1-\frac{\hat{s}_{ij,2}^{e_{i}e_{j}}+D_{ij}}{l^{e_i}}  \right )t_{i}^{e_{i,1}}+\frac{\hat{s}_{ij,2}^{e_{i}e_{j}}+D_{ij}}{l^{e_i}}t_{i}^{e_{i,2}} \le \left ( 1-\theta_{ij,2}^{e_{i}e_{j}}  \right )t_{j}^{e_{j,1}}\\
&+\theta_{ij,2}^{e_{i}e_{j}} t_{j}^{e_{j,2}}+M\left(1-\delta _{ij}^{e_i}\right) ,\\
&\forall i,j \in \mathcal{N},Angle\left(e_i,e_j \right) \le 90^{\circ}.
\end{aligned}
\end{equation}

\begin{equation}
\begin{aligned}
\label{C_C_T2}
&\left ( 1-\frac{\hat{s}_{ij,1}^{e_{i}e_{j}}+D_{ij}}{l^{e_i}}  \right )t_{i}^{e_{i,1}}+\frac{\hat{s}_{ij,1}^{e_{i}e_{j}}+D_{ij}}{l^{e_i}}t_{i}^{e_{i,2}} \ge \left ( 1-\theta_{ij,2}^{e_{i}e_{j}}  
 \right )t_{j}^{e_{j,1}}\\
&+\theta_{ij,2}^{e_{i}e_{j}} 
 t_{j}^{e_{j,2}}-M\delta _{ij}^{e_i},\\
& \left ( 1-\frac{\hat{s}_{ij,2}^{e_{i}e_{j}}+D_{ij}}{l^{e_i}}  \right )t_{i}^{e_{i,1}}+\frac{\hat{s}_{ij,2}^{e_{i}e_{j}}+D_{ij}}{l^{e_i}}t_{i}^{e_{i,2}} \le \left ( 1-\theta_{ij,1}^{e_{i}e_{j}}  \right )t_{j}^{e_{j,1}}\\
&+\theta_{ij,1}^{e_{i}e_{j}} t_{j}^{e_{j,2}}+M\left(1-\delta _{ij}^{e_i}\right),\\
&\forall i,j \in \mathcal{N},Angle\left(e_i,e_j \right) \ge 90^{\circ}.
\end{aligned}
\end{equation}
Note that $Angle\left(e_i,e_j \right)$ denotes the angle between $e_i$ and $e_j$. The collision avoidance constraints are linear constraints with respect to $t_i^{e_{i,1}}$ and $t_i^{e_{i,2}}$.
\subsection{MILP Formulation}
The objective function should account for important indicators such as arrival time, velocity changes during travel, acceleration changes, and turning effects to adapt to the requirements of complex traffic situations and improve operational efficiency and passenger comfort.

For CAV $i$, the part of the objective function related to arrival time can be represented as:
\begin{equation}
\label{f_t}
f_{i,t}=\sum_{v\in des_i}t_{i}^{v}.
\end{equation}

The part of the objective function related to velocity changes can be represented as:
\begin{equation}
\label{f_V}
f_{i,V}=\sum_{e\in \mathcal{E} _i}\left ( \overline{s}_{i}^{e}+\underline{s}_{i}^{e} \right ). 
\end{equation}

With the term $f_{i, V}$ as a penalty term, the velocity deviation of CAV $i$ is penalized, leading to the velocity of CAV $i$ tracking the reference velocity. The part of the objective function related to longitudinal acceleration can be represented as:
\begin{equation}
\label{f_a}
\begin{aligned}
f_{i,a} & = \sum_{v\in \mathcal{\tilde{V}}_{i}}\sum_{\alpha \in\mathcal{E}_{i}^{v,in}}\sum_{\beta \in\mathcal{E}_{i}^{v,out}}\sum_{k \in\mathcal{K}}\left ( V_{i,k}^{r} \right )^2\left ( \overline{\gamma}_{i,v}^{\alpha\beta,k}+\underline{\gamma}_{i,v}^{\alpha\beta,k} \right ) \\
& +\sum_{\beta \in\mathcal{E}_{s_i}^{v,out}}\sum_{k \in\mathcal{K}}\left ( V_{i,k}^{r} \right )^2\left ( \overline{\gamma}_{i,s_i}^{\beta,k}+\underline{\gamma}_{i,s_i}^{\beta,k} \right ). 
\end{aligned}
\end{equation}

With the part $f_{i, a}$ as a penalty term, the velocity changes of CAV $i$ are constrained, leading to the improvement of passengers comfort. The part of the objective function related to steering can be represented as:
\begin{equation}
\label{f_theta}
\begin{aligned}
f_{i,\theta}&=\sum_{v\in \mathcal{\tilde{V}}_{i}}\sum_{\alpha \in\mathcal{E}_{i}^{v,in}}\sum_{\beta \in\mathcal{E}_{i}^{v,out}}\sum_{k \in\mathcal{K}}\eta_{i,v}^{\alpha\beta,k}\\ &+\sum_{\beta \in\mathcal{E}_{s_i}^{v,out}}\sum_{k \in\mathcal{K}}\eta_{i,s_i}^{\beta,k}.
\end{aligned}
\end{equation}

With the part $f_{i,\theta}$ as a penalty term, lateral accelerations induced by steering are penalized, leading to the generated paths without unnecessary lane changes and abrupt turns. In conclusion, with all the penalty terms, for each CAV $i$, the objective function can be represented as:
\begin{equation}
J_{i}=\alpha _tf_{i,t}+\alpha _Vf_{i,V}+\alpha _af_{i,a}+\alpha _\theta f_{i,\theta }
\end{equation}
where $\alpha _t$, $\alpha _V$, $\alpha _a$, and $\alpha _\theta$ are the weight parameters of each objective function part. With the constraints, the decision-making model of each CAV $i$ can be defined as:
\begin{equation}
\begin{aligned}
\label{model_1}
    \min &\quad\alpha _tf_{i,t}+\alpha _Vf_{i,V}+\alpha _af_{i,a}+\alpha _\theta f_{i,\theta }\\
    \text{s.t.} & \quad
    (\ref{Path_c_1})-(\ref{Ro_c_2}),(\ref{C_C_T1})-(\ref{C_C_T2}).
\end{aligned}
\end{equation}

We introduce $c$ as the number of constraints and $n$ as the number of all the CAVs variables. For $x_i$, which we define as $x_i=[t_{i,v},\overline{s}_{i}^{e},....,\eta_{i,v}^{\alpha\beta,k}]^T$, it contains all the variables of CAV $i$. With the variables of the surrounding CAVs, we define $x=\left( x_i,x_{-i} \right)\in \mathbb{R}^{n}$ as the vector of all the decision variables in the neighborhood $\mathcal{N}_{i}$, where $x_{-i}$ represents the variables of surrounding CAVs. Finally, we transform the (\ref{model_1}) into a compact form:
\begin{equation}
\begin{aligned} 
\label{model_2} 
 \underset{x_{i}}{\text{min}} & && J_i\left( x_i \right) \\ 
 \text{s.t.} & && Ax\leq b 
\end{aligned} 
\end{equation}
for some suitable $A\in \mathbb{R}^{ c\times n } $ and $b\in \mathbb{R}^{n}$. For $i\in \mathcal{N}$, we define $X=\left [ x_1,x_2,...,x_N \right ]^T$. Due to objective function and all of the constraints are linear, the decision-making model is a standard MILP problem.
\section{Mixed-Integer Potential Game Formulation}
In principle, by calculating the optimal solution for (\ref{model_2}), we can obtain the decision variables for each CAV and the reference times of passing through each node, thereby determining the paths and time profiles of each CAV. For the decision-making model (\ref{model_2}), for each CAV, solving the decision-making model independently yields the optimal decision for each CAV. However, the collision constraints between CAVs are coupled, and independent solutions cannot guarantee that all collision constraints are satisfied. Moreover, the cooperative solution is NP-hard\cite{bixby2004mixed} with a large problem scope. As the number of CAVs increases, the computational time for the optimal solution increases exponentially, making it impractical for real-time applications. Therefore, to tackle the problem of computational time growing exponentially as the scale of the problem and to ensure the safety of the decision strategy, we transform the MILP problem into an MIPG.

For each CAV, we define the feasible sets $\mathcal{X}_{i}\left( x_{-i}\right)=\left \{ x_i\in \mathbb{R}^{n_i}\mid A\left ( x_i,x_{-i} \right )\le b   \right \}$ and $\mathcal{X}=\left \{ x\in \mathbb{R}^{n}\mid Ax\le b   \right \}$. Moreover, we introduce a function $P:\mathbb{R}^{n}\to \mathbb{R}$, which we define as $P\left ( x \right )=\sum_{i \in \mathcal{N}}J_i \left ( x_i \right )$. For all $i\in \mathcal{N}$, and for all $x_i,y_i\in \mathcal{X}_i\left ( x_{-i} \right )$, (\ref{P_func}) holds true.
\begin{equation}
\label{P_func}
P\left ( x_i,x_{-i} \right )-P\left ( y_i,x_{-i} \right )=J_i \left ( x_i \right )-J_i \left ( y_i \right ).
\end{equation}
By \cite{facchinei2011decomposition}, $P$ is an exact potential function for the proposed problem. Given the strategies $x_{-i}$ of surrounding CAVs for CAV $i$, the best strategies of CAV $i$ is:
\begin{equation}
\label{model_game}
\begin{aligned} 
\underset{x_{i}}{\text{argmin}}& && J_i\left( x_i \right) \\ 
\text{s.t.} & && \left( x_i,x_{-i} \right) \in \mathcal{X}.
\end{aligned} 
\end{equation}
Then we define the notion of equilibrium solution.
\begin{definition}
    If there exists a positive real number $\varepsilon$ and $x^*\in \mathcal{X}$ such that 
    \begin{equation}
J_i\left ( x_i^* \right )\le \underset{x_{i}\in \mathcal{X}_i\left ( x_{-i}^* \right )}{\text{inf}}J_i \left ( x_i \right )+\varepsilon
\end{equation}
    holds for all $i\in \mathcal{N}$. We called $x^*\in \mathcal{X}$ is an $\varepsilon$-Mixed-Integer Nash Equilibrium ($\varepsilon$-MINE) of the game in (\ref{model_game}).
\end{definition}
All $x^*$ constitute the $\varepsilon$-approximate minimal set. With the above definition, the $\varepsilon$-approximate minimal set on $\mathcal{X}$ is a subset of the $\varepsilon$-MINE strategy \cite{sagratella2017algorithms}, which means for the $\varepsilon$-approximate minimal value set $x \in \mathcal{X}$, any $x^* \in \mathcal{X}$ such that $P\left( x^* \right) \leq P\left( x \right) + \varepsilon$ is an $\varepsilon$-MINE of the game. Informally, an $\varepsilon$-MINE is a set of driving strategies that are almost individually optimal, given the safety constraints.

\section{Gauss-Seidel Algorithm for Decision-Making}
As is well known, solving MIPG is a challenging task, due to integer variables involved. Here, we propose algorithms to compute the $\varepsilon$-MINE for the game problem through an iterative process. $x_i\left ( k \right )$ and $x_{-i}\left ( k \right ),k\in \mathbb{N}$ represent vectors of decision variables of the CAV $i$ and its surrounding CAVs in the $k$-th iteration, respectively. we introduce the cost variation $\triangle J_i\left ( k \right )$ at one iteration:
\begin{equation}
\triangle J_i\left ( k \right ) =J_i\left ( x_i\left (k \right ) \right )-J_i\left ( \check{x}_i\left (k \right ) \right )
\end{equation}
where $\check{x}_i\left ( x_{-i}\left (k \right ) \right ) \in x_i^*\left (k \right )$.
\subsection{Gauss-Seidel Algorithm for MIPG}
We first propose a Gauss-Seidel algorithm for cooperative decision-making . This algorithm solves an optimization problem in each iteration process, i.e., solves the optimization problem for a specific CAV separately, obtains the optimal decision variables $\check{x}_i$ for the CAV in the current situation, and compares them with the original decision to get the decrease in cost. If the decrease in cost is smaller than $\varepsilon$, the original decision $x_i\left (k \right )$ is retained; otherwise, the optimal decision $\check{x}_i\left (k \right )$ is adopted. To sum up, the Gauss-Seidel algorithm for cooperative decision-making is outlined in Algorithm \ref{alg1}.
\begin{algorithm}[t]
\caption{Gauss-Seidel Algorithm for Cooperative Decision-Making}
\label{alg1}
\begin{algorithmic}[1]
    \STATE Choose an initial state $x\left (0 \right )\in \mathcal{X}$, set $k=0$, $sign=0$
    \WHILE {$sign\ne1$} 
    \FOR{$i = 1$ to $N$}
    \STATE Solve the MILP problem of (\ref{model_2}) with $x\left (k \right )$ and get a solution of $\check{x}_i\left ( x_{-i}\left (k \right ) \right )$
    \STATE $\triangle J_i\left ( k \right ) =J_i\left ( x_i\left (k \right ) \right )-J_i\left (  \check{x}_i\left (k \right ) \right )$
    \STATE $x_i\left (k+1 \right )=
    \begin{cases}
     \check{x}_i\left (k \right ),&\text{if $\triangle J_i \ge \varepsilon$} \\
     x_i\left (k \right ),&\text{else}
    \end{cases}
    $
    \STATE $x\left (k+1 \right )=\left (x_i\left (k+1 \right ), x_{-i}\left (k \right ) \right )$
    \ENDFOR
    \IF{$x\left (k+1 \right )=x\left (k \right )$}
    \STATE $sign=1$
    \ENDIF
    \STATE $k=k+1$
    \ENDWHILE
\end{algorithmic}
\end{algorithm}

To demonstrate that the decision vector computed by algorithm is feasible, namely, $x^*\in \mathcal{X}$ of the MIPG, we first suppose $\left (x_i\left (k \right ),x_{-i}\left (k \right )\right )$ is feasible, i.e., $\left (x_i\left (k \right ),x_{-i}\left (k \right )\right ) \in \mathcal{X}$. The claim of $\left (x_i\left (k+1 \right ),x_{-i}\left (k \right )\right )$  is true if the original decision is retained, i.e., $x_i\left (k+1 \right )=x_i\left (k \right )$. On the other hand, if the decision of CAV $i$ changes, i.e., $x_i\left (k+1 \right )\in \check{x}_i\left ( x_{-i}\left (k \right ) \right )$, we obtain that $x_i\left (k+1 \right )\in \mathcal{X}_{i}\left( x_{-i}\left (k \right )\right)$ and  $\left (x_i\left (k+1 \right ),(x_{-i}\left (k \right )\right ) \in \mathcal{X}$. Notice that $x\left (0 \right )\in \mathcal{X}$, the decision vector computed by the algorithm is feasible.
\subsection{Sequential Gauss-Seidel Algorithm for MIPG}
When multiple CAVs travel in the same direction on the single lane, certain general cases present no collision risk. For instance, there is no collision risk when the rear CAV's velocity is significantly slower than the front CAV's velocity or when the rear CAV's velocity matches the front CAV's velocity but with consistently lower longitudinal acceleration. In these scenarios, there is no necessity for the rear CAV to change lanes or overtake. Due to the relatively low collision risk in these situations, they are not our primary concern. To ensure the algorithm's applicability in all situations, we focus on the special case where the rear CAV's velocity is substantially higher than the front CAV's velocity.

\begin{figure}[t]
    \centering
    \subfloat[An example of an unsolvable situation. The decision variables of black and blue CAV are fixed while optimizing for the white CAV. The white CAV can not overtake the front CAVs because both roads are congested and it cannot change the decision variables of the front CAVs in its own iteration. The collision is occurred.]{\includegraphics[width=0.48\textwidth]{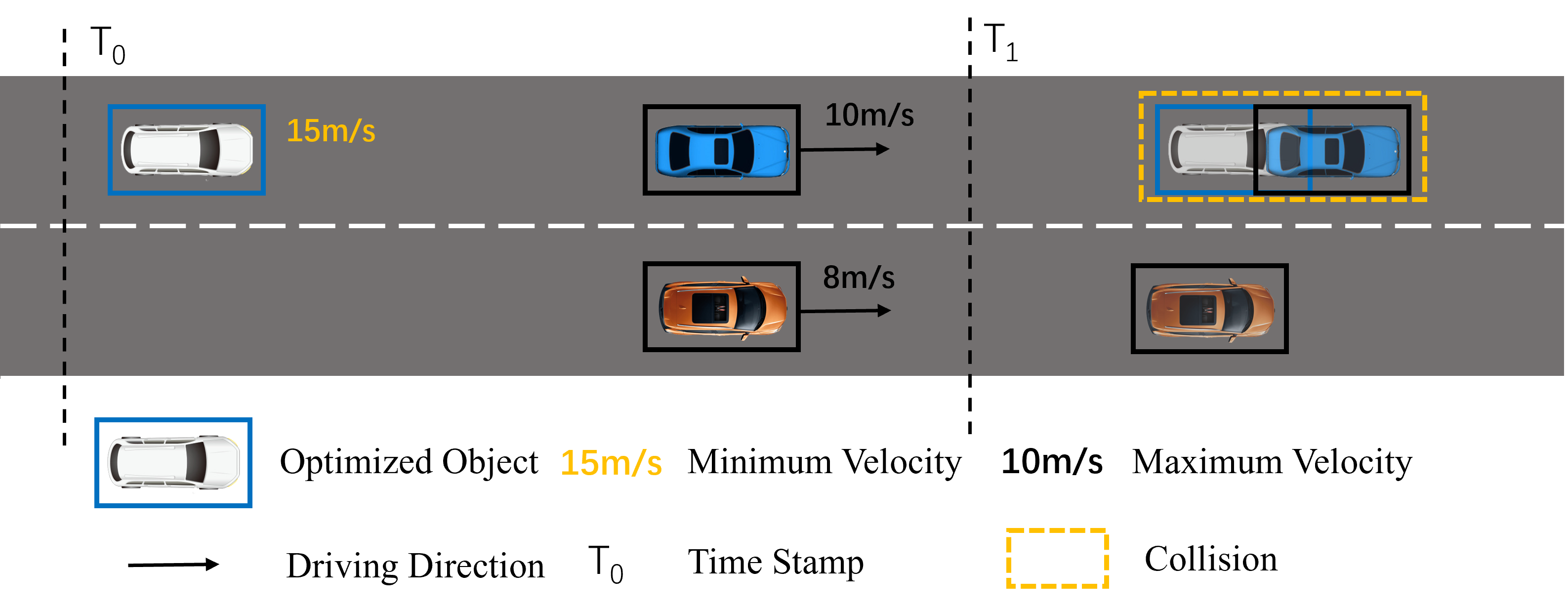}\label{non-solution}}
    \vspace{-3mm}
    \subfloat[An example of a different optimization sequence. By optimizing the blue CAV first, the blue CAV chooses to change lanes to overtake the black CAV to avoid collision with the white CAV, which allows the white CAV to continue straight without collision. Thus, the problem becomes solvable.]{\includegraphics[width=0.48\textwidth]{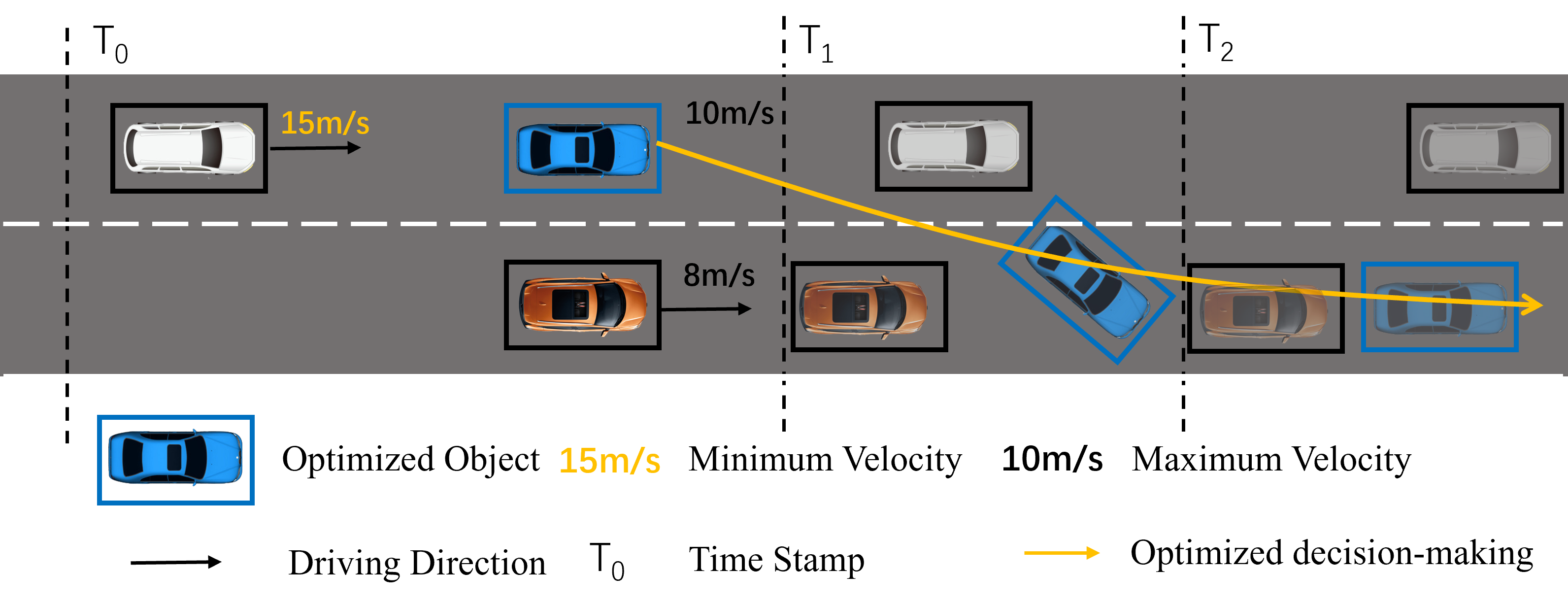}\label{good-solution}}
    \caption{Examples of different optimization sequence.}
    \label{Sequential examples}
\end{figure}

In Algorithm \ref{alg1}, optimization is performed sequentially based on the order of the CAVs' positions in the direction of travel, starting from CAVs with an initial position at the back to those with an initial position at the front. This optimization sequence makes it challenging for CAVs in intermediate positions with slower velocities to find suitable driving strategies, potentially leading to sub-optimal solutions, or even in unsolvable situations extreme cases as illustrated in Fig. \ref{Sequential examples}\subref{non-solution}. Furthermore, when optimization is based solely on positional order, collisions in rear CAVs may render the optimization problem unsolvable. Therefore, employing this method imposes strict requirements on the feasibility of the initial state variables. However, the different optimization sequences result in different final solutions. Since each CAV cannot alter the driving strategy of other CAVs during individual optimization, it can only choose to yield or overtake to avoid collision. Therefore, different optimization sequences result in different decisions for CAVs, as illustrated in Fig. \ref{Sequential examples}\subref{good-solution}. The appropriate optimization sequence is crucial for the efficiency and quality of the solution.

To address the issue of unsolvable optimization problems in extreme situations and to remove the strict limitations on the feasibility of the initial decision variables, we propose an improved algorithm based on Algorithm \ref{alg1}. We evaluate the optimization priority for each CAV and provide an optimization ranking to improve the efficiency and quality of the optimization.

Firstly, we introduce the definition of the optimization priority for the CAVs. For CAV $i$, its optimization sequence $O_i$ is obtained by the function of the reference factors $w$ and the corresponding weights $\beta$, i.e., $O_i=f\left ( w,\beta \right ) $, where $w= [ w_1,w_2,...,w_m  ]^T $  and $\beta=[ \beta_1,\beta_2,...,\beta_m ]^T$. In this problem, the main reference factors are the initial position $\left ( p_{i}^{x,ini},p_{i}^{y,ini} \right )$ and initial velocity $V_{i}^{ini}$. The initial position reflects the complexity of the CAV interactions. When CAVs are positioned further back in the direction of travel, there are more important edge pairs and interactions become more complex. When prioritizing the optimization of CAVs with complex interactions before handling other CAVs, changes in the other CAVs can easily cause the initial CAV's decision to change again, while the decisions of other CAVs with simpler interactions are less likely to change significantly due to collision constraints. Additionally, optimizing decisions for CAVs with complex interactions requires more time. Therefore, CAVs located closer to the front should be optimized first to avoid repetitive processing of CAVs with complex interactions. The initial velocity reflects the flexibility of the CAV's adjustment strategy. The CAVs with slower velocity take more time to change lane, making it difficult to perform rapid yielding and overtaking maneuvers. Therefore, the optimization sequence should prioritize CAVs with slower initial speeds so that the strategies of other CAVs can be based on the strategies of the slower CAVs. We propose some methods to obtain the optimization sequence by considering both factors.
\subsubsection{Linearized Optimization Order Determination Method}
In the linearized optimization order determination (LOD) method, we define the final ranking as the ranking of the linearly superimposed value of the initial position ranking and the initial velocity ranking, which can be expressed as:
\begin{equation}
Order=Sort\left (\beta_{p}Sort_p\left (p^{ini} \right )+\beta_{V}Sort_V\left ( V^{ini}\right ) \right )
\end{equation}
where the $Sort_p$ function outputs the initial positions in the order of distance from far to near in the direction of travel, the $Sort_V$ function outputs the initial velocities in increasing order, and the $Sort$ function outputs the function values in increasing order. 
\subsubsection{Technique for Order Preference by Similarity to Ideal Solution Method}
In LOD method, the rankings of two factors are linearly combined, which inevitably leads to the generation of identical values. When ranking identical values, it is only possible to follow the original order and further differentiation is not possible, thus affecting the final result. To avoid this issue, a method using technique for order preference by similarity to ideal solution method (TOPSIS) has been proposed, which fully utilizes the comprehensive evaluation of original data to accurately reflect the differences between evaluation schemes. This method is mainly divided into the following three steps:
\begin{itemize}
\item Step 1: Positive direction transformation of the original matrix.
\item Step 2: Normalization of positive matrix.
\item Step 3: Calculate scores.
\end{itemize}

The purpose of Step 1 is to normalize the original indices, i.e., to convert the indices into maximization type indices, so that all indices are aligned in the same direction. In this problem, we introduce the initial data matrix $\mathcal{M}$ as $\mathcal{M}=\left [ p^{ini},V^{ini} \right ]$, where $p^{ini}=\left [ p^{ini}_1,p^{ini}_2,...,p^{ini}_N \right ]^T $ and $V^{ini}=\left [ V_{1}^{ini},V_{2}^{ini},...,V_{N}^{ini} \right ]^T$. As aforementioned, when determining the optimization sequence, we first optimize CAVs that are located closer to the front and have slower velocities. Therefore, the indicator after normalization can be represented as:
    \begin{equation}
    \begin{aligned}
        & \bar{p}^{ini} =p^{ini}, \\
        & \bar{V}_i^{ini} = \max \left (V^{ini}\right )-V_i^{ini} , \\
        & \bar{\mathcal{M}}=\left [ \bar{p}^{ini},\bar{V}^{ini} \right ].
    \end{aligned}
    \end{equation}

After normalization, the requirements for priority optimization are satisfied better if the value of $\bar{p}^{ini}$ and $\bar{V}_i^{ini}$ are larger. However, the units of the two values are currently different. In order to eliminate the units influence, Step 2 normalizes the positive matrix. We define the element in the $i$-th row and $j$-th column of the positive direction transformation matrix $\bar{\mathcal{M}}$ as $m_{ij}$, and the normalization method is represented as:
    \begin{equation}
    \check{m}_{ij}=\frac{m_{ij}-\min\left ( \bar{\mathcal{M}}_j  \right ) }{\max\left ( \bar{\mathcal{M}}_j  \right )-\min\left ( \bar{\mathcal{M}}_j  \right )} 
    \end{equation}
where $\bar{\mathcal{M}}_j$ represents the elements of the $j$-th column of the positive direction transformation matrix $\bar{\mathcal{M}}$. 

In Step 3, we calculate the distance between the normalized positive matrix and the maximum and minimum values to obtain scores. The scores are used for sorting based on their magnitude. We define its distance from the maximum and minimum values as $D_i^+$ and $D_i^-$, respectively, which can be represented as:
        \begin{equation}
        \begin{aligned}
        & D_i^+ = \sqrt{\beta _j\left ( \max\left ( \bar{\mathcal{M}}_j  \right )-\check{m}_{ij} \right )^2 }, \\
        & D_i^- = \sqrt{\beta _j\left ( \check{m}_{ij}-\min\left ( \bar{\mathcal{M}}_j  \right )\right )^2 }
        \end{aligned}
        \end{equation}
        where $\beta _j$ is the weight value corresponding to the $j$-th column. With $D_i^+$ and $D_i^-$, the score can be calculated as:
        \begin{equation}
        \mathcal{S}_i=\frac{D_i^-}{D_i^++D_i^-}.
        \end{equation}
        The final priority can be obtained by arranging the score $\mathcal{S}_i$ in descending order. 
\begin{algorithm}[t]
\caption{Linearized and Enhanced Optimization Order Determination Algorithm}
\label{alg3}
\begin{algorithmic}[1]
\renewcommand{\algorithmicrequire}{\textbf{Input:}}
\renewcommand{\algorithmicensure}{\textbf{Output:}}
    \REQUIRE Initial Position $p^{ini}=\left [ p^{ini}_1,p^{ini}_2,...,p^{ini}_N \right ]^T $, Initial Velocity $V^{ini}=\left [ V_{1}^{ini},V_{2}^{ini},...,V_{N}^{ini} \right ]^T,\beta_p , \beta_V $\\
    \ENSURE  Order of Optimization \\
    \STATE $Order_{p^{ini}}=Sort_{p}\left (p^{ini} \right )$
    \STATE $Order_{V^{ini}}=Sort_V\left (V^{ini} \right )$
    \FOR{$i = 1$ to $N$}
    \STATE $Order_{i,{p}}=find\left ( Order_{{p^{ini}}}==i\right )$
    \STATE $Order_{i,V}=find\left ( Order_{V^{ini}}==i\right )$
    \STATE $Value_i=\beta_{p}Order_{i,{p^{ini}}}+\beta_{V}Order_{i,V^{ini}}$
    \ENDFOR
    \STATE $Order_{LOD}=Sort\left (Value \right )$
    \STATE $\bar{p}^{ini} =p^{ini}$
    \FOR{$i = 1$ to $N$}
    \STATE $\bar{V}_i^{ini} = \max \left (V^{ini}\right )-V_i^{ini}$
    \ENDFOR
    \STATE $\bar{\mathcal{M}}=\left [ \bar{p}^{ini},\bar{V}^{ini} \right ]$
    \STATE $\beta_1=\beta_p,\beta_2=\beta_V$
    \FOR{$i = 1$ to $N$}
    \FOR{$j = 1$ to $2$}
    \STATE   $\check{m}_{ij}\!=\!\left (m_{ij}\!-\!\min\left ( \bar{\mathcal{M}}_j  \right )\right ) \!/\! \left (\max\!\left ( \bar{\mathcal{M}}_j  \right )\!-\!\min\left ( \bar{\mathcal{M}}_j  \right )\right )$ 
    \STATE $ D_i^+ = \sqrt{\beta _j\left ( \max\left ( \bar{\mathcal{M}}_j  \right )-\check{m}_{ij} \right )^2 }$
    \STATE  $D_i^- = \sqrt{\beta _j\left ( \check{m}_{ij}-\min\left ( \bar{\mathcal{M}}_j  \right )\right )^2 }$
    \STATE  $\mathcal{S}_i=\frac{D_i^-}{D_i^++D_i^-}$
    \ENDFOR
    \ENDFOR
    \STATE $Order_{TOPSIS}=$ Sort $\mathcal{S}_i$ in descending order
\end{algorithmic}
\end{algorithm}

The LOD and TOPSIS algorithms are shown in Algorithm \ref{alg3}. To remove the strict limitations on the feasibility of the initial decision variables, we propose an algorithm to detect the number of collisions each CAV generates with other CAVs in the initial state and determine an optimization order according to this. Due to the randomness of the initial state, it is difficult for each CAV to avoid collisions with other CAVs. Therefore, in each iteration, we sort the optimization order of CAVs $Order_{Col}$ based on the number of collisions. CAVs with a higher number of collisions have more complex interaction relationships, requiring more time for optimization and are easily influenced by decisions of other CAVs. Therefore, CAVs with fewer collisions are prioritized for optimization to avoid repeated optimization of CAVs with higher collision counts. 
When the number of collisions is $0$ for all CAVs during the iteration, it indicates that the current state is a feasible solution. In this case, the optimization can be performed based on the $Order$ obtained using the two methods mentioned above. 


With these techniques, the sequential Gauss-Seidel algorithm for cooperative decision-making is shown in Algorithm \ref{alg4}.
\begin{algorithm}[t]
\caption{Sequential Gauss-Seidel Algorithm for Cooperative Decision-Making}
\label{alg4}
\begin{algorithmic}[1]
    \STATE Randomly choose an initial state $x\!\left (0 \right)\!$, set $k\!=\!0$, $sign\!=\!0$\!
    \STATE Obtain the $Order$ using Algorithm \ref{alg3}
    \WHILE {$sign\ne1$} 
    \STATE Sort the order $Order_{Col}$ according to the number of collisions using (\ref{C_C_T1}) and (\ref{C_C_T2})
    \FOR{all $i \in Order_{Col}$}
    \STATE Solve the MILP problem of (\ref{model_2}) with $x\left (k \right )$ and get a solution of $\check{x}_i\left ( x_{-i}\left (k \right ) \right )$
    \STATE $\triangle J_i\left ( k \right ) =J_i\left ( x_i\left (k \right ) \right )-J_i\left (  \check{x}_i\left (k \right ) \right )$
    \STATE $x_i\left (k+1 \right )=
    \begin{cases}
     \check{x}_i\left (k \right ),&\text{if $\triangle J_i \ge \varepsilon$} \\
     x_i\left (k \right ),&\text{else}
    \end{cases}
    $
    \STATE $x\left (k+1 \right )=\left (x_i\left (k+1 \right ), x_{-i}\left (k \right ) \right )$
    \ENDFOR
    \IF{$x\left (k+1 \right )=x\left (k \right )$}
    \STATE $sign=1$
    \ENDIF
    \STATE $k=k+1$
    \ENDWHILE
\end{algorithmic}
\end{algorithm}

\subsection{Trajectory Planning}
In the previous sections, we get the vector of decision variables. The paths and the time related to the nodes can serve as good references for generating collision-free and feasible trajectories for every CAV. In this section, we regard time variables $t_{i}^v$ and path variables $P_{i}^e$ computed by the methods we proposed in previous section as references, and formulate an optimal control problem to generate a trajectory for every CAV. The kinematic and control input constraints are the main considerations in this process.
\subsubsection{Kinematic Constraints}
For all CAVs, we adopt a discrete-time bicycle motion model with a constant time interval $\tau_s$. In the constraints, we assume that the initial time steps of all CAVs are aligned with each other, with an initial time $t_i^{s_i}=0$. However, the duration of the generated CAV reference points is different, so to ensure that the generated trajectories have the same duration, we can use the shortest trajectory as a standard to trim the other trajectories. Therefore, the initial time steps of all CAVs in the next planning period are still consistent.

We denote the reference path of CAV $i$ obtained from solving methods as $\left \{ v_i^k \right \} $, where $v_i^k$ is the $k$-th vertex that CAV $i$ passes through in the path, and $k \in \left \{ 0,1,...,m^i \right \} $. For simplicity, we denote the time required for CAV $i$ to reach $v_i^k$ from the center as $t_i^k$, and the corresponding position of $v_i^k$ as $p_i^k$. Furthermore, we assume $t_i^0=0$ for all $i \in \mathcal{N}$, with the first time step being $0$. Then the last time step of CAV $i$ is $\tau_i=\left \lfloor t_{i}^{m_i} /  \tau_{s}   \right \rfloor    $, where $\left \lfloor \cdot \right \rfloor  $ represents the floor function. The set of all time steps for CAV $i$ is denoted as $T_i=\left [0,1,..., \tau_i\right ]$.

The state vector $x_i^\tau$, $\tau \in T_i$ is defined as $x_i^\tau=\left [p_{i,x}^{\tau},p_{i,y}^{\tau},\theta_{i}^{\tau},V_{i}^{\tau}  \right ]^T$. $p_{i,x}^{\tau}$ and $p_{i,y}^{\tau}$ represent the $X$ and $Y$ coordinates of the CAV's rear axle center in the global coordinate system, while $\theta_{i}^{\tau}$ represents the heading angle, and $V_{i}^{\tau}$ represents the longitudinal velocity. The control inputs are denoted as $u_{i}^{\tau}=\left[ \delta_{i}^{\tau} ,a_{i}^{\tau} \right ]^T $, where $\delta_{i}^{\tau}$ is the steering angle, and $a_{i}^{\tau}$ is the acceleration. With above definition, the discrete-time bicycle motion model \cite{7995816} can be expressed as:
\begin{equation}
\begin{cases}
\label{TP_c_1}
     p_{i,x}^{\tau+1}=p_{i,x}^{\tau}+f_r\left (\delta_{i}^{\tau} ,V_{i}^{\tau} \right  ) \cos \left(\theta_{i}^{\tau} \right) ,\\
     p_{i,y}^{\tau+1}=p_{i,y}^{\tau}+f_r\left (\delta_{i}^{\tau} ,V_{i}^{\tau} \right  ) \sin \left(\theta_{i}^{\tau} \right) ,\\
     \theta_{i}^{\tau+1}=\theta_{i}^{\tau}+\arcsin \left(\tau_s V_{i}^{\tau}\sin\left(\delta_{i}^{\tau} \right) / b_i \right) ,\\
     V_{i}^{\tau+1}=V_{i}^{\tau}+\tau_s a_{i}^{\tau}
\end{cases}
\end{equation}
where $b_i$ is the wheelbase of the CAV $i$ and $f_r\left (\delta_{i}^{\tau} ,v_{i}^{\tau} \right  )$ is the function defined as:
\begin{equation}
    f_r\left (\delta ,V \right  )=b+\tau_s V \cos{\delta}-\sqrt{b^2-\left (\tau_s V \sin\left ( \delta  \right )  \right )^2 }. 
\end{equation}
\subsubsection{Control Input Constraints}
Due to physical limitations on engine force, braking force, and wheel angles, the following constraints need to be applied to the control inputs:
\begin{equation}
\begin{aligned}
\label{TP_c_2}
& \delta_{i}^{min} \le \delta_{i}^{\tau} \le \delta_{i}^{max}, \\
& a_{i}^{min} \le a_{i}^{\tau} \le a_{i}^{max}, \\
& \forall i \in \mathcal{N},\forall \tau \in T_i.
\end{aligned}
\end{equation}
\subsubsection{Reference Path}
We use linear interpolation to generate the reference point positions at each time step. Specifically, the corresponding reference point positions are:
\begin{equation}
\begin{aligned}
& \hat{p}_i^{\tau,ref}=\frac{\tau\times \tau_s-t_i^k}{t_i^{k+1}-t_i^k} p_i^{k+1}+\frac{t_i^{k+1}-\tau\times \tau_s}{t_i^{k+1}-t_i^k} p_i^{k}, \\
& \forall k \in \left \{ 0,1,...,m_i \right \},t_i^k\le \tau\times \tau_s \le t_i^{k+1}, \\
& \forall i \in \mathcal{N},\forall \tau \in T_i.
\end{aligned}
\end{equation}
$\hat{p}_i^{\tau,ref}$ is the position of the center point of CAV $i$, but in the constraints of the motion model, the position of the point refers to the position of the rear axle center of the CAV. Therefore, the position of the reference point needs to be moved and transformed to the rear axle center. Thus, the derived position after the transformation is:
\begin{equation}
\begin{aligned}
& p_i^{\tau,ref}=\hat{p}_i^{\tau,ref}-d_i\frac{p_i^{k+1}-p_i^{k}}{\left \| p_i^{k+1}-p_i^{k} \right \| _2}, \\
& \forall k \in \left \{ 0,1,...,m_i \right \},t_i^k\le \tau\times \tau_s \le t_i^{k+1}, \\
& \forall i \in \mathcal{N},\forall \tau \in T_i
\end{aligned}
\end{equation}
where $d_i$ is the distance between the center point and the rear axle center point and $\left \| \ \right \|  _2$ represents the 2-norm.
\subsubsection{Optimal Control Formulation}
With above definition, we formulate the optimal control problem for each CAV as :
\begin{equation}
\begin{aligned}
\label{TP_model}
 \underset{u_{i}^{\tau}}{\text{min}} & \sum_{\tau=0}^{\tau_i} \left(x_i^\tau-x_i^{\tau,ref}\right)^TQ\left(x_i^\tau-x_i^{\tau,ref}\right)+\sum_{\tau=0}^{\tau_i-1}\left(u_i^\tau\right)^TRu_i^\tau, \\
 \text{s.t.} &  \quad (\ref{TP_c_1}),(\ref{TP_c_2}).
\end{aligned}
\end{equation}
The objective function takes the deviations of actual trajectories from the reference and control input into consideration. In particular, $x_i^{\tau,ref}=\left [ p_i^{\tau,x,ref},p_i^{\tau,y,ref},\theta_i^{\tau,ref},V_i^{\tau,ref} \right ]^T$, where $p_i^{\tau,x,ref}$ and $p_i^{\tau,y,ref}$ represent the $X$ and $Y$ coordinates of reference paths in the global coordinate system, respectively, $\theta_i^{\tau,ref}$ is the reference heading angle and $V_i^{\tau,ref}$ is the reference velocity. $Q$ is positive semi-definite and $R$ is positive definite. For each CAV $i$, solving (\ref{TP_model}) can obtain a feasible and collision-free trajectory.
\section{Experiment Results}
For verifying the effectiveness of the proposed method, three common urban traffic scenarios are designed in this section. Gurobi\cite{gurobi} and CasADi\cite{Andersson2018} are used to solve (\ref{model_1}) and (\ref{TP_model}), respectively. The relevant parameter settings are shown in Table \ref{tab_1}. The matrices $Q$ and $R$ in the problem are set as:
$Q=diag(20,20,0,0)$ and $R=diag(20,0.1)$. 

For each scenario, we have set up three types of experiments. The first type of experiment visually demonstrates the algorithm's effectiveness in reducing the total cost and the impact of each constraint in a single trial. The convergence is reflected by the average cost change per iteration. The second type of experiment is a quantitative study. We conduct 25 sets of experiments  using different methods we proposed with random initial states, where we add a Gauss distribution to the initial velocities $V^{ini}$. This can be expressed as $\tilde{V}^{ini}=V^{ini}+\lambda $, 
where $\lambda \sim N(0,3^2)$. We evaluate the algorithm efficiency of the proposed algorithm by the time to solve the $\varepsilon$-MINE and the average time $T_{Aver}$ per iteration. In the third type of experiment, we conduct an additional 10 sets of experiments using the methods we proposed and MILP method, with the same experimental setup as the second group of experiments. All the experiments are conducted with an Intel Core i7-9750H processor in MATLAB.

\begin{table}[t]
\centering
\caption{Parameter Settings Used in Simulations}
\label{tab_1}
\begin{tabular}{@{}l@{\hspace{12pt}}c@{\hspace{12pt}}c@{\hspace{12pt}}c@{\hspace{12pt}}@{\hspace{12pt}}c@{\hspace{12pt}}c@{}}
\toprule 
\quad Param. & Value & Param. & Value & Param. & Value \\ \midrule 
\quad $\gamma_{max}$ & $3.0$\,$\textup{m/s}^2$ & $\gamma_{min}$ & $-4.5$\,$\textup{m/s}^2$ & $\eta_{max}$ & $3.0$\,$\textup{m/s}^2$ \\
\quad  $\alpha_{t}$ & $0.1$ & $\alpha_{V}$ & $1.0$ & $\alpha_{a}$ & $0.5$ \\ 
\quad  $\alpha_{\theta}$ & $0.5$ & $\tau_s$ & $0.1$\,$\textup{s}$ & $b$ & $2.405$\,$\textup{m}$ \\
\quad  $d_{safe}$ & $2.366$\,$\textup{m}$ &  $L$ & $3.526$\,$\textup{m}$ & $W$ & $1.673$\,$\textup{m}$ \\
\quad  $a_{max}$ & $4.0$\,$\textup{m/s}^2$ & $a_{min}$ & $-6.0$\,$\textup{m/s}^2$ & $\epsilon$ & $0.2$\\
\quad $\delta_{max}$ & $0.9$\,$\textup{rad}$ &  $\delta_{min}$ & $-0.9$\,$\textup{rad}$ & $\beta_p,\beta_V$ & $0.5$ \\ 
\bottomrule 
\end{tabular}
\end{table}

\subsection{Overtaking}
\begin{figure*}[htbp]
     \centering
     \subfloat[Way-Point Graph]{\includegraphics[width=0.33\linewidth]{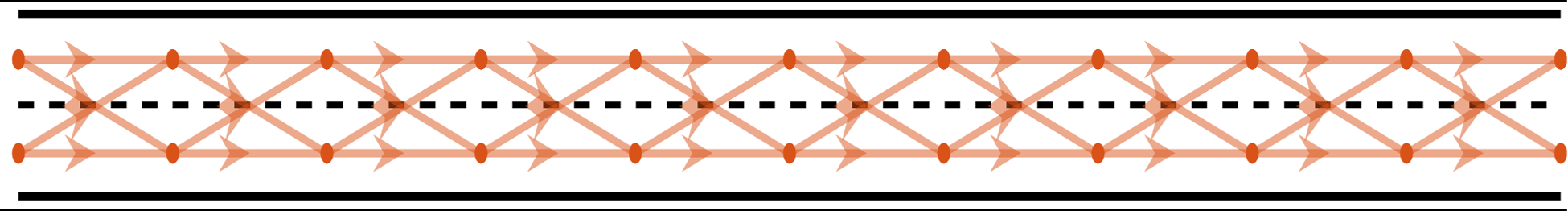}
     \label{Stra_Waypoint}
     }
    \subfloat[Initial Position]{\includegraphics[width=0.33\linewidth]{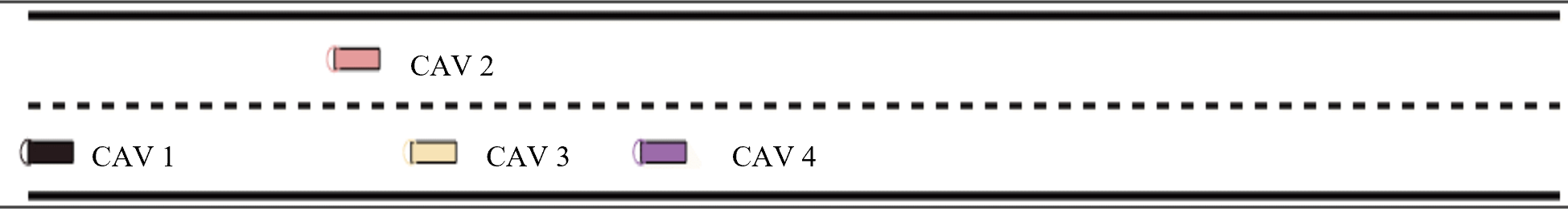}
     \label{Stra_Initial}
     }
    \subfloat[$t=0.5$\,s]{\includegraphics[width=0.33\linewidth]{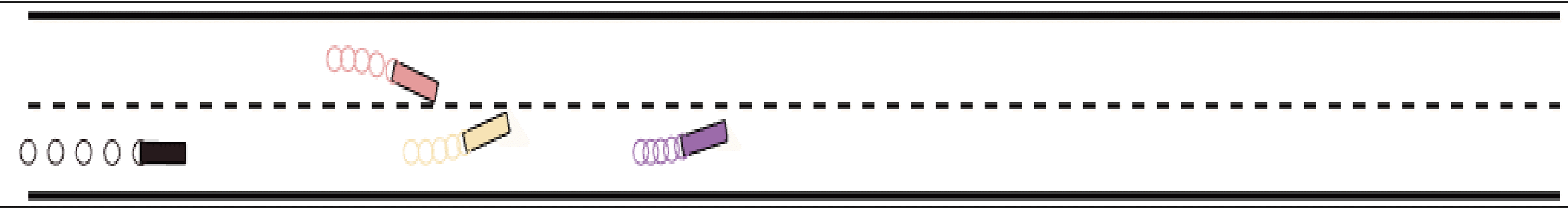}
     \label{Stra_ini_0.5}
     }
    \vspace{-3mm}
     \subfloat[$t=3.6$\,s]{\includegraphics[width=0.33\linewidth]{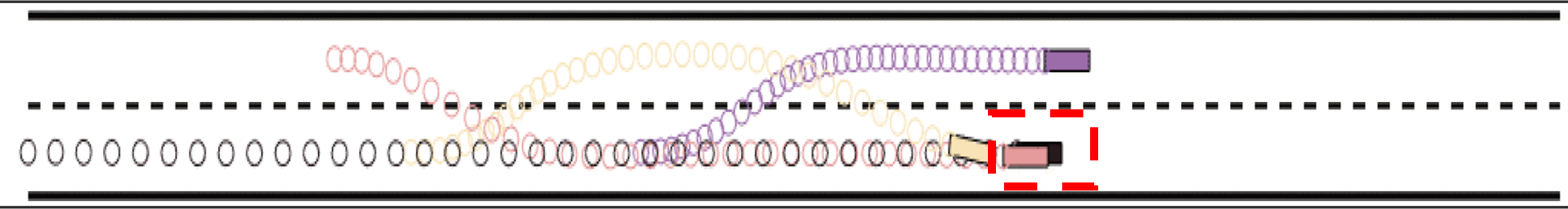}
     \label{Stra_ini_3.6}
     }
     \subfloat[$t=5.4$\,s]{\includegraphics[width=0.33\linewidth]{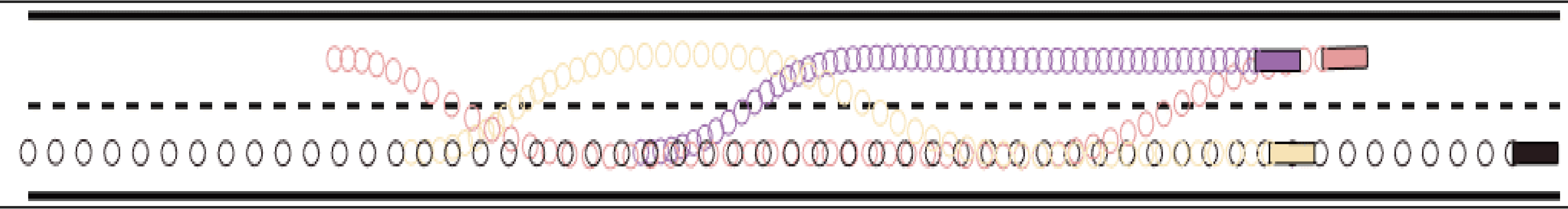}
     \label{Stra_ini_5.4}
     }
     \subfloat[$t=0.8$\,s]{\includegraphics[width=0.33\linewidth]{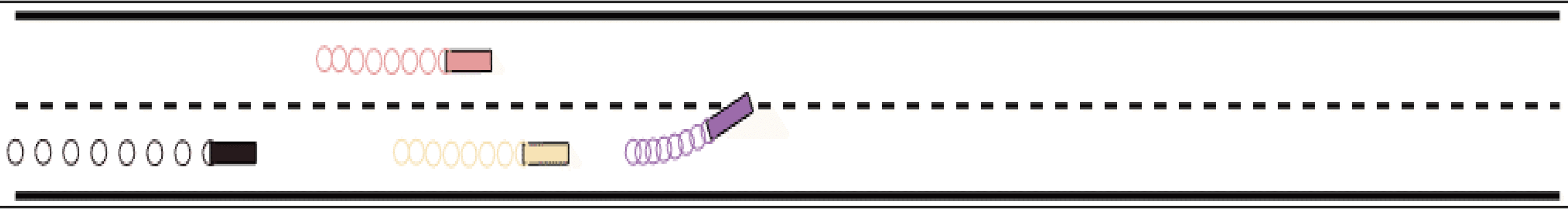}
     \label{Stra_0.8}
     }
    \vspace{-3mm}
    \subfloat[$t=3.6$\,s]{\includegraphics[width=0.33\linewidth]{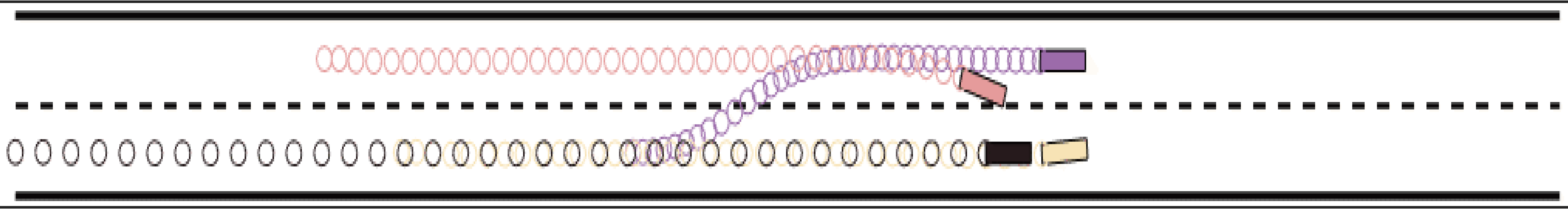}
     \label{Stra_3.6}
     }
         \subfloat[$t=4.6$\,s]{\includegraphics[width=0.33\linewidth]{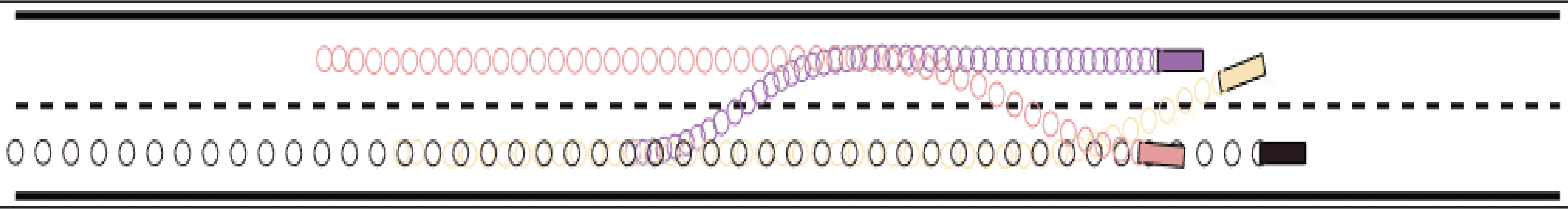}
     \label{Stra_4.6}
     }
    \subfloat[$t=5.4$\,s]{\includegraphics[width=0.33\linewidth]{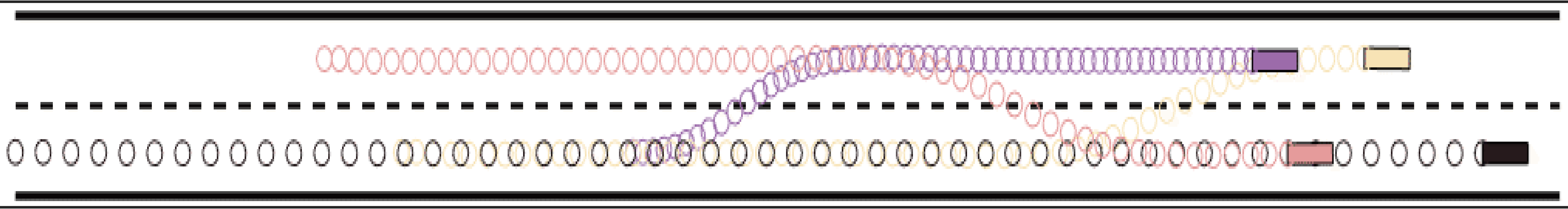}
     \label{Stra_5.4}
     }
    \caption{Simulation results for the overtaking scenario.}
    \label{Stra}
\end{figure*}

In this section, we demonstrate examples of CAVs overtaking on a one-way two-lane road. We sample along the centerline of the lane with a spacing of $10$\,m. Each sampled waypoint is connected to the two nearest waypoints on the adjacent lane in the direction of travel to form edges. Each lane is $3.75$\,m wide. The plotted way-point graph is shown in the Fig. \ref{Stra}\subref{Stra_Waypoint} and the initial positions of each CAV are shown in the Fig. \ref{Stra}\subref{Stra_Initial}. The initial speed $V^{ini}$ of CAV1-4 are $18$\,m/s, $12$\,m/s, $12$\,m/s, and $8$\,m/s, respectively, and the reference velocity $V_i^{r}$ are set as the same as the initial speed with distribution $\tilde{V}^{ini}$. The velocity range for all CAVs are defined as $\left[0.6\tilde{V}^{ini},1.3\tilde{V}^{ini} \right]$, which represents the minimum velocity $V^{min}$ and maximum velocity $V^{max}$, respectively. 

\begin{figure}[t]
    \includegraphics[width=0.95\linewidth]{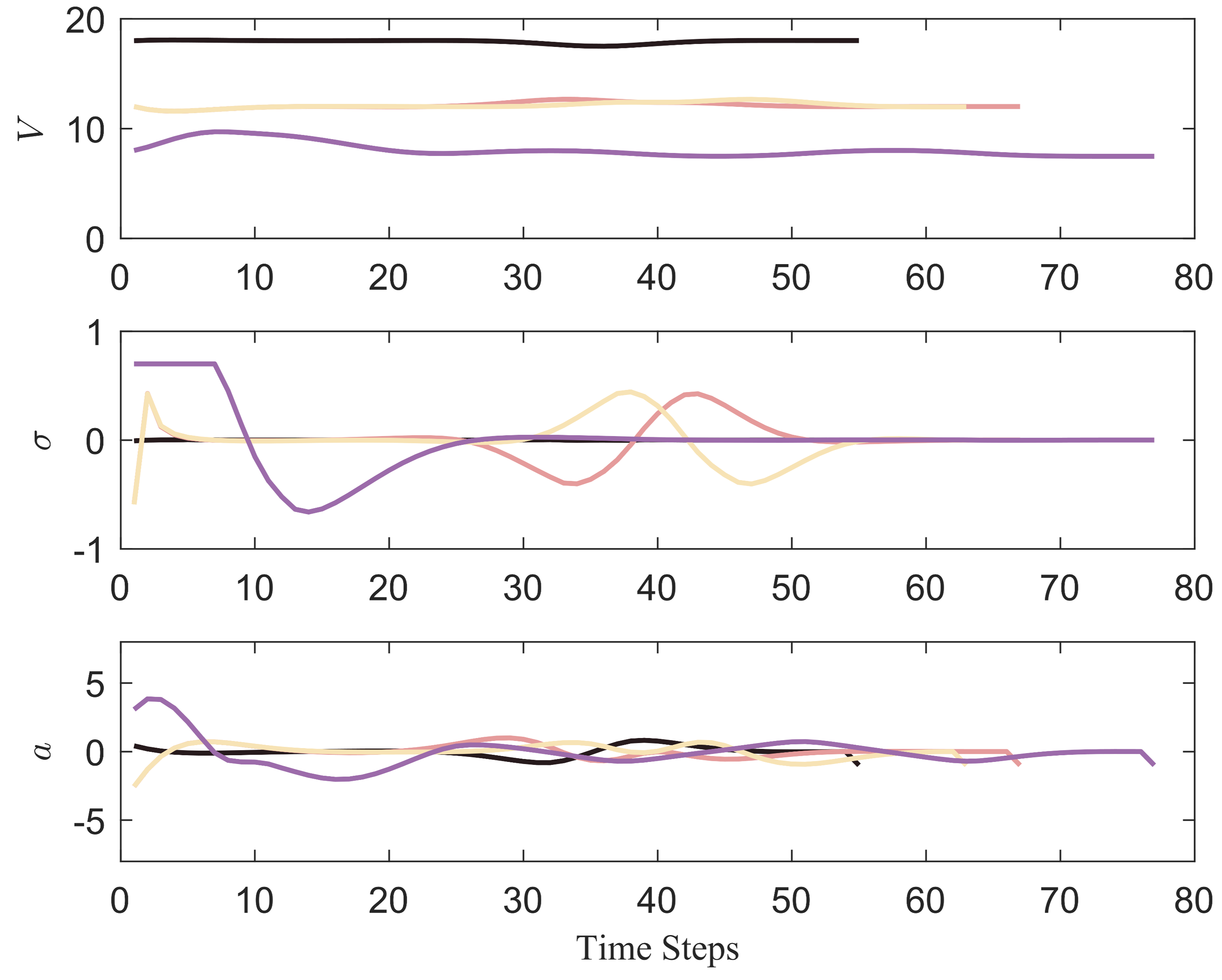}
    \caption{Velocity, acceleration, and steering angle of all CAVs for the overtaking scenario.}
    \label{Stra-para2} 
\end{figure}

In the first experiment, it is evident that the speed of the rear CAV is much greater than that of the front CAV, and CAV2 and CAV3 occupy two lanes. Therefore, to achieve collision-free safe driving, a coordinated strategy is crucial. To demonstrate the effectiveness of the proposed algorithm, we randomly assign initial states, with the initial decision strategies as shown in the Fig. \ref{Stra}\subref{Stra_ini_0.5}-\subref{Stra_ini_5.4}. It can be observed that in the initial decision strategies, a collision between CAV1 and CAV3 occurs at $t=3.6$\,s, and throughout the driving process, CAVs exhibit unreasonable lane-changing behavior. Results are shown in Fig. \ref{Stra}\subref{Stra_0.8}-\subref{Stra_5.4}. Unreasonable lane-changing behaviors, as exhibited by CAV2 and CAV3 at $t=0.5$\,s, have been modified to more reasonable actions, such as maintaining a straight trajectory. Potential collisions are averted by CAV3 adjusting lanes after advancing a certain distance and CAV2 changing lanes only after CAV1 has safely passed. Since CAV4 has the slowest speed, both CAV2 and CAV3 use the lane-changing overtaking method to overtake CAV4 and avoid collisions. The circles of different colors represent the trajectories of the corresponding CAVs. The average cost decreased from the initial value of $7.89$ to $3.93$ after $4$ iterations, resulting in a $50\%$ reduction in average cost. Fig. \ref{Stra-para2}  shows the changes in speed, acceleration, and steering wheel angle.

\begin{table}
\centering
\caption{Comparison of computing time and cost value among different methods in different optimization orders for the overtaking scenario.
}
\label{Stra_time}
\begin{tabular}{|c|c|c|c|}
\hline
Methods & TOPSIS Order & Linear Order & Default Order \\
\hline
Computing Time &$2.37$\,s & $2.41$\,s & $2.87$\,s \\
\hline
$T_{Aver}$ & $0.57$\,s & $0.58$\,s & $0.59$\,s \\
\hline
$J_{Aver}$ & $2.0$ & $2.0$ & $2.7$ \\
\hline
Success Rate   & $22/25$ & $22/25$ & $22/25$ \\
\hline
\end{tabular}
\end{table}

 The results of the second experiment are shown in Table \ref{Stra_time}, where default order refers to the sequential arrangement of CAVs from left to right according to their initial positions, as shown in the Fig. \ref{Stra}\subref{Stra_Initial}. $T_{Aver}$ represents the computational time per iteration and $J_{Aver}$ represents the final cost per CAV. We represent this optimization sequence using the array $\left[1,2,3,4\right]$, where the numbers correspond to the CAV indices arranged in the order of optimization from left to right.

\begin{figure*}[t]
    \centering
     \subfloat[Way-point Graph]{\includegraphics[width=0.17\linewidth,height=4.3cm]{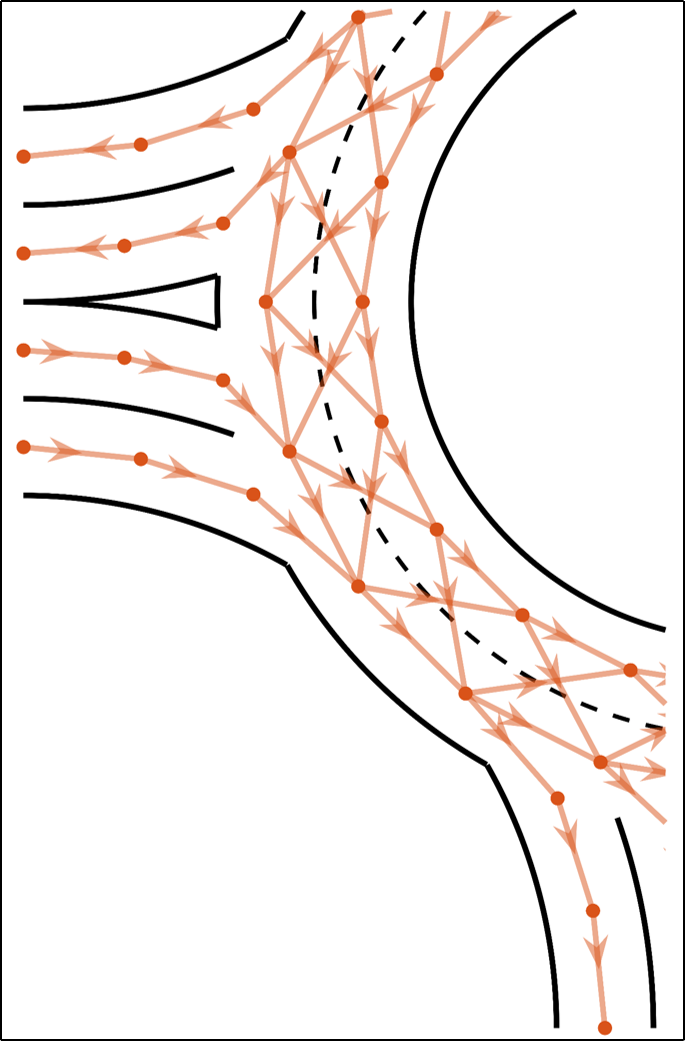}
     \label{Round_Waypoint}
     }
     \hspace{1mm}
    \subfloat[$t=0.9$\,s]{\includegraphics[width=0.17\linewidth,height=4.3cm]{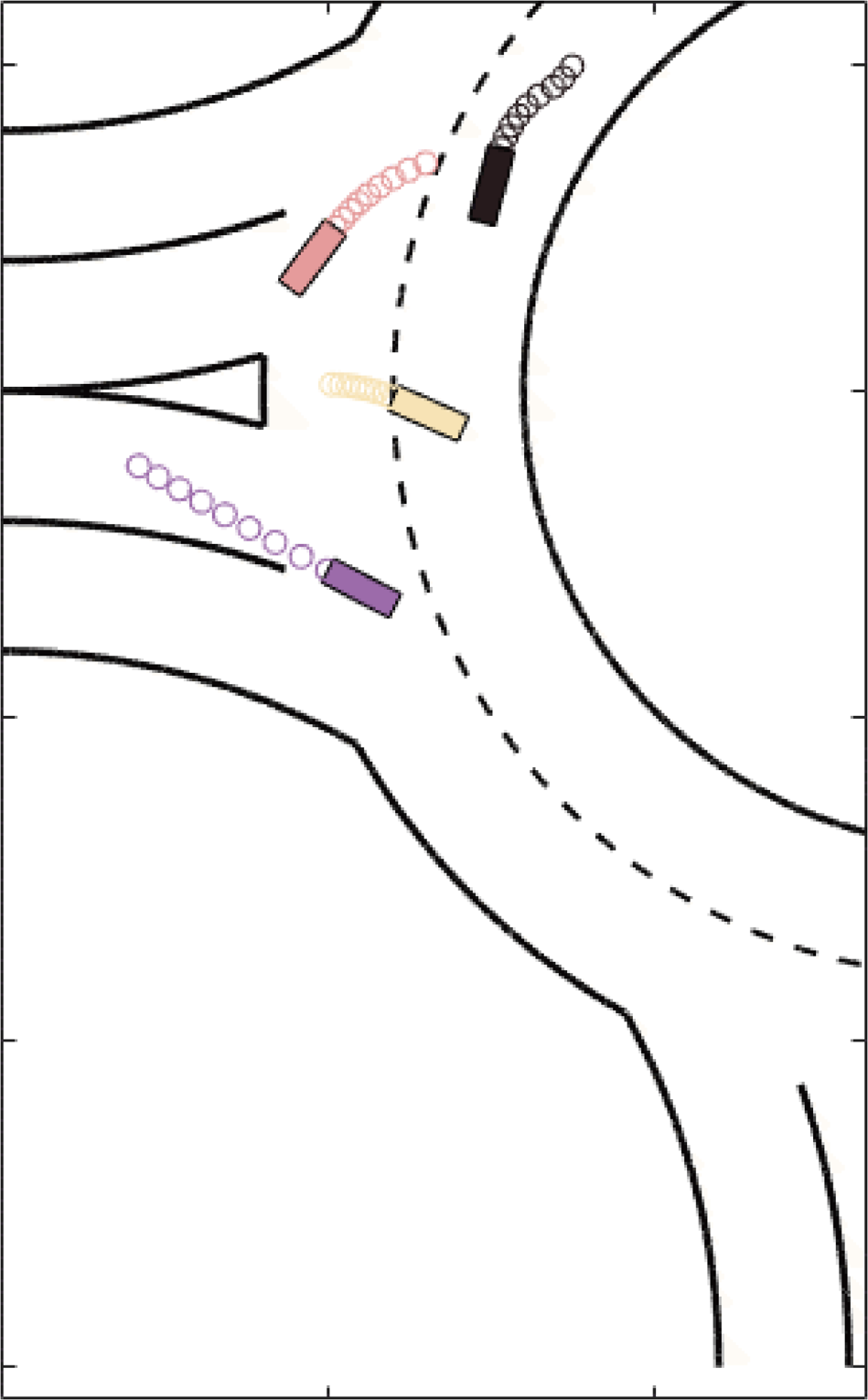}
     \label{Round_ini_0.9}
     }
      \hspace{1mm}
         \subfloat[$t=3.5$\,s]{\includegraphics[width=0.17\linewidth,height=4.3cm]{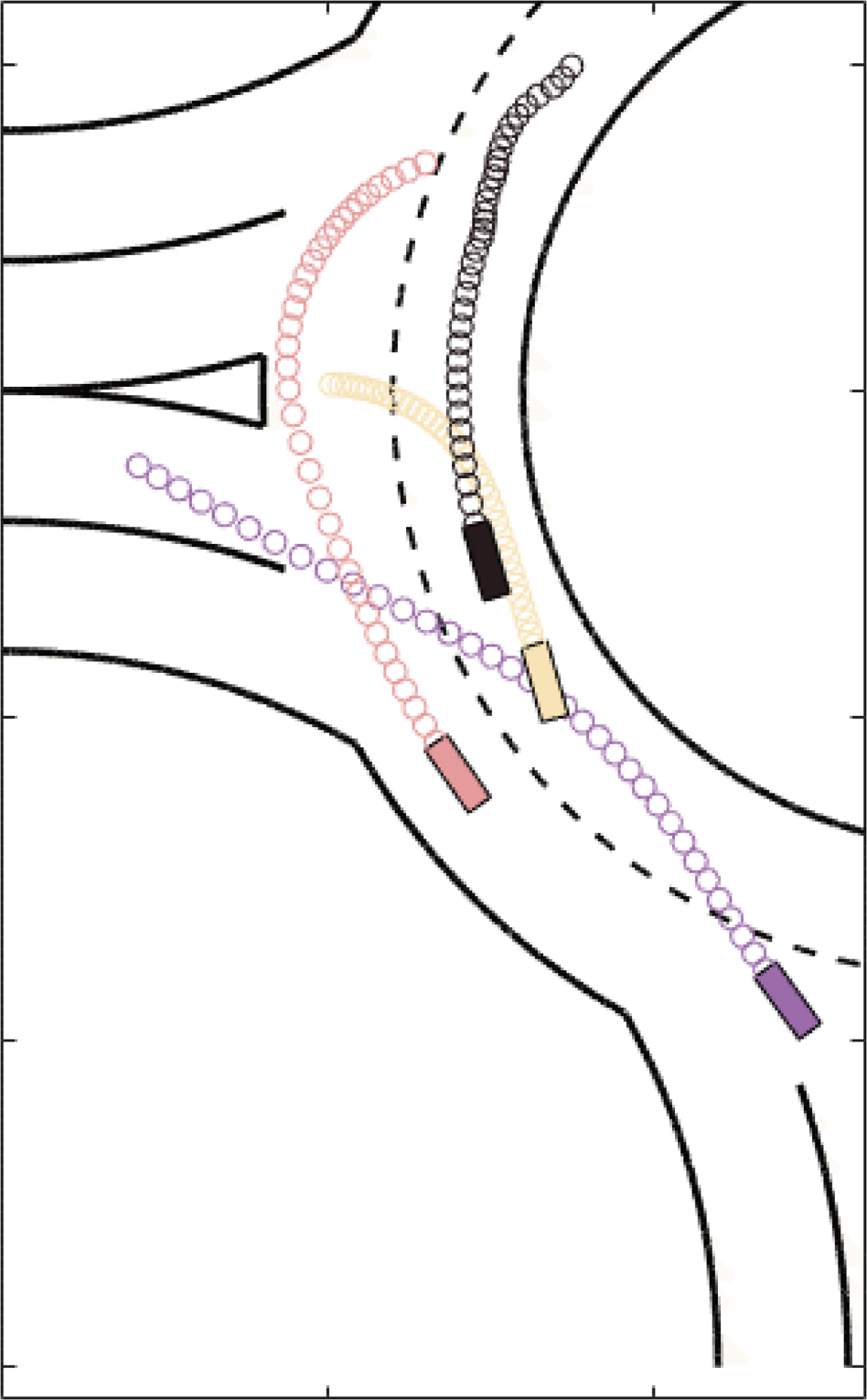}
     \label{Round_ini_3.5}
     }
      \hspace{1mm}
         \subfloat[$t=0.9$\,s]{\includegraphics[width=0.17\linewidth,height=4.3cm]{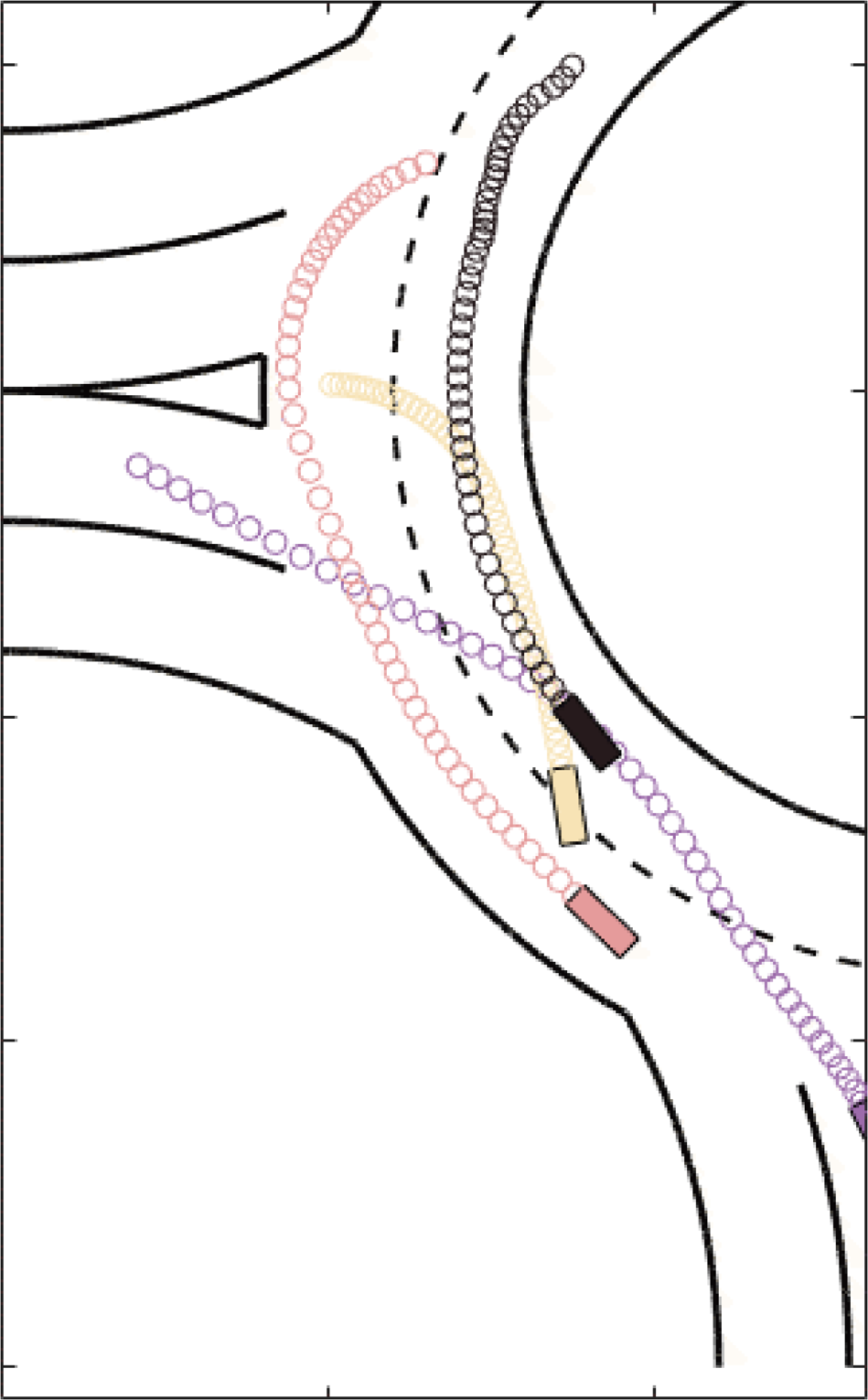}
     \label{Round_ini_4.5}
     }
      \hspace{1mm}
         \subfloat[$t=7.0$\,s]{\includegraphics[width=0.17\linewidth,height=4.3cm]{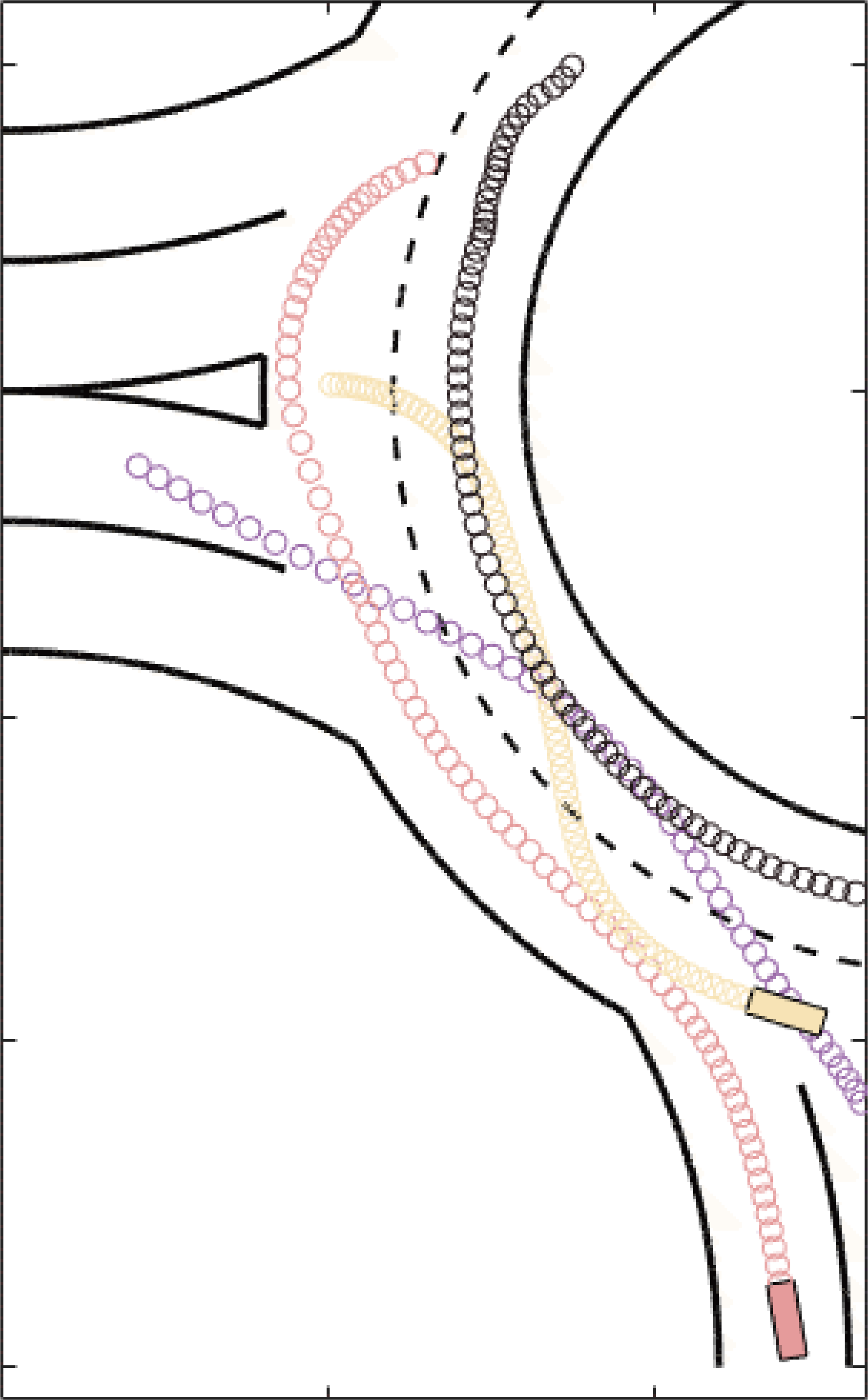}
     \label{Round_ini_7}
     }
     \vspace{-3mm}
         \subfloat[Initial Position]{\includegraphics[width=0.17\linewidth,height=4.3cm]{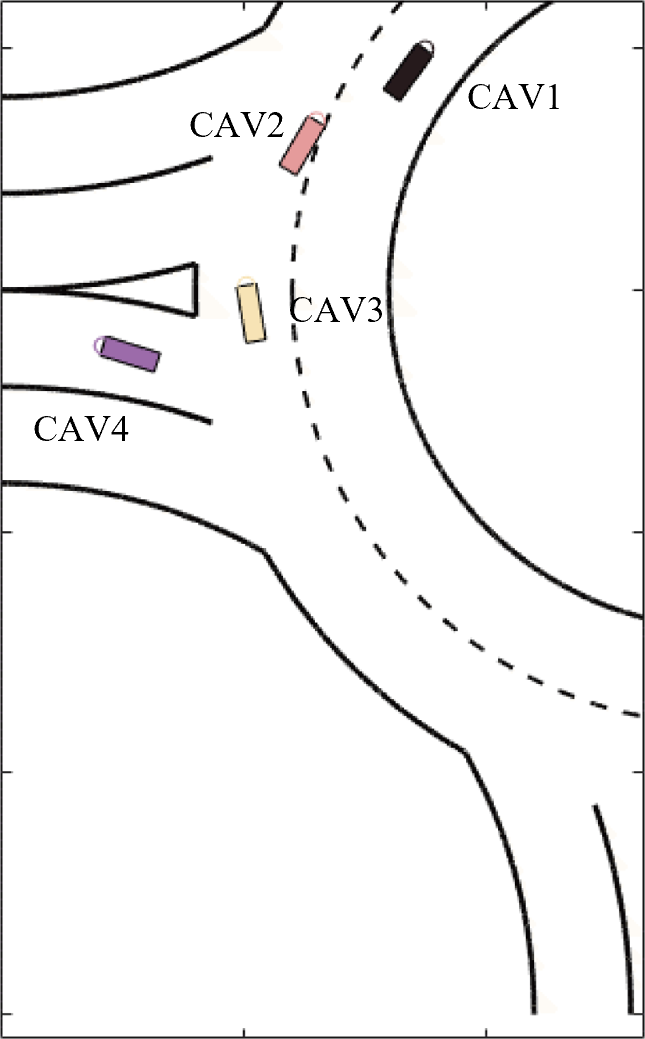}
     \label{Round_Initial}
     } \hspace{1mm}	
    \subfloat[$t=1.1$\,s]{\includegraphics[width=0.17\linewidth,height=4.3cm]{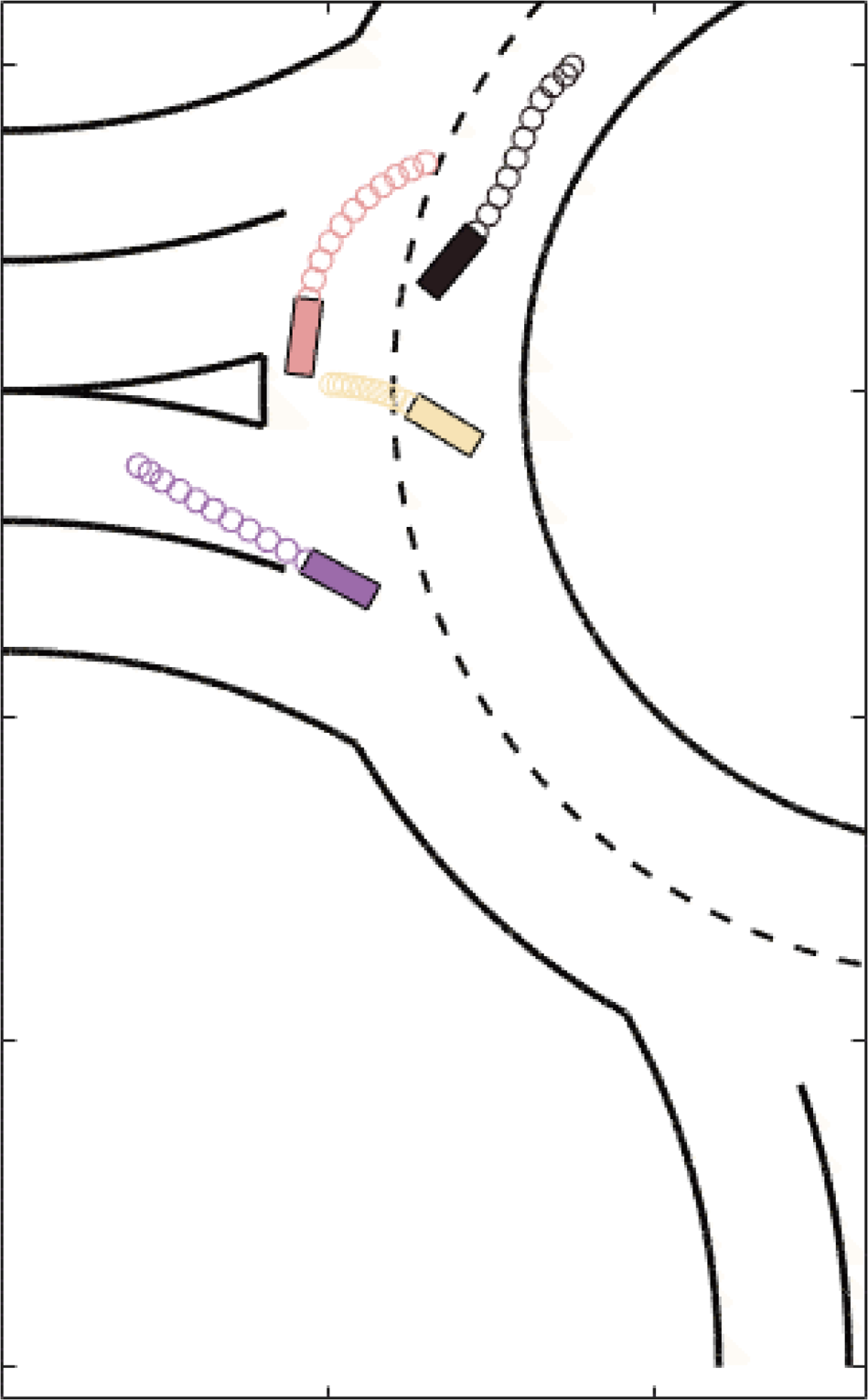}
     \label{Round_1.1}
     } \hspace{1mm}
         \subfloat[$t=2.4$\,s]{\includegraphics[width=0.17\linewidth,height=4.3cm]{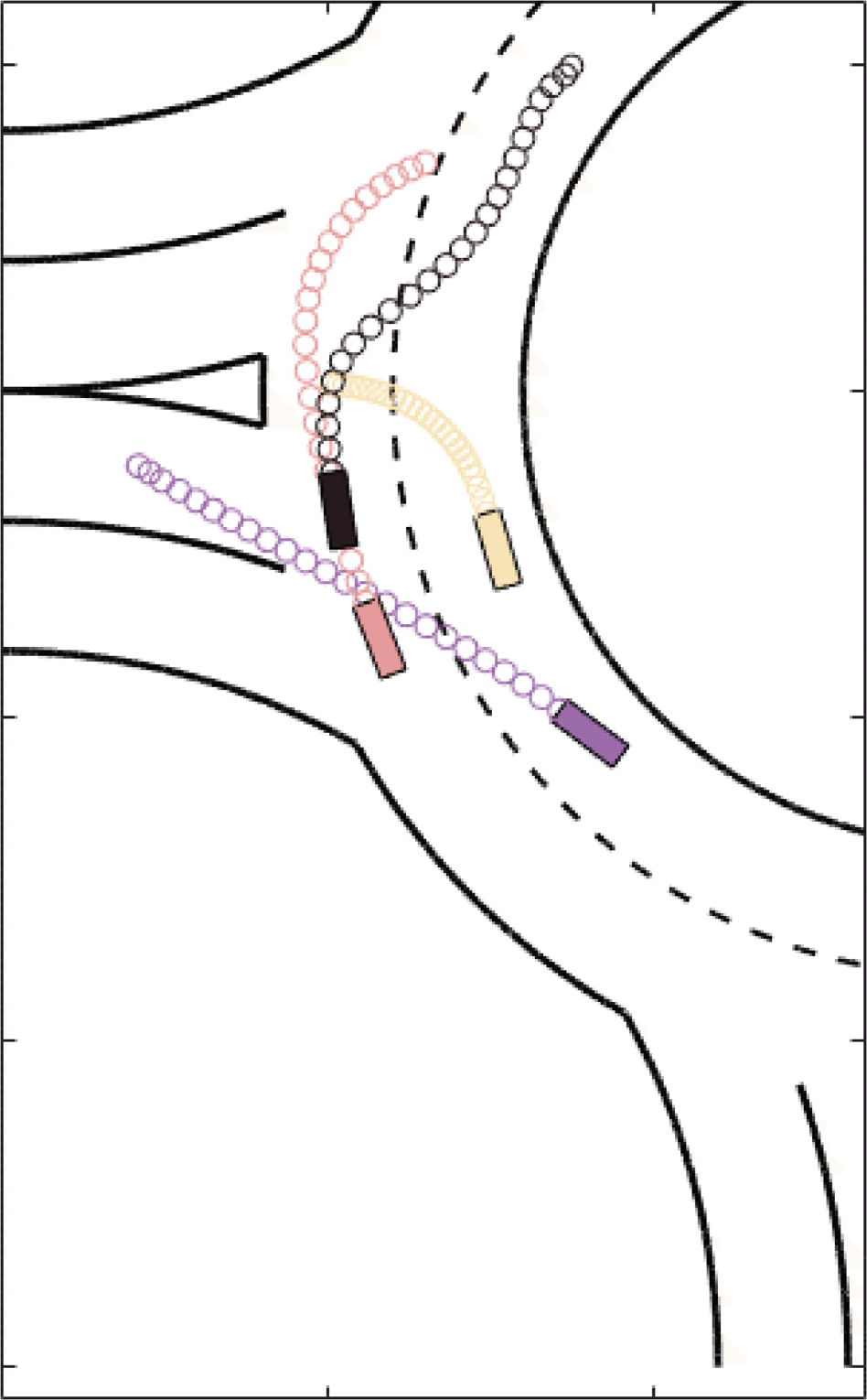}
     \label{Round_2.4}
     } \hspace{1mm}
         \subfloat[$t=3.6$\,s]{\includegraphics[width=0.17\linewidth,height=4.3cm]{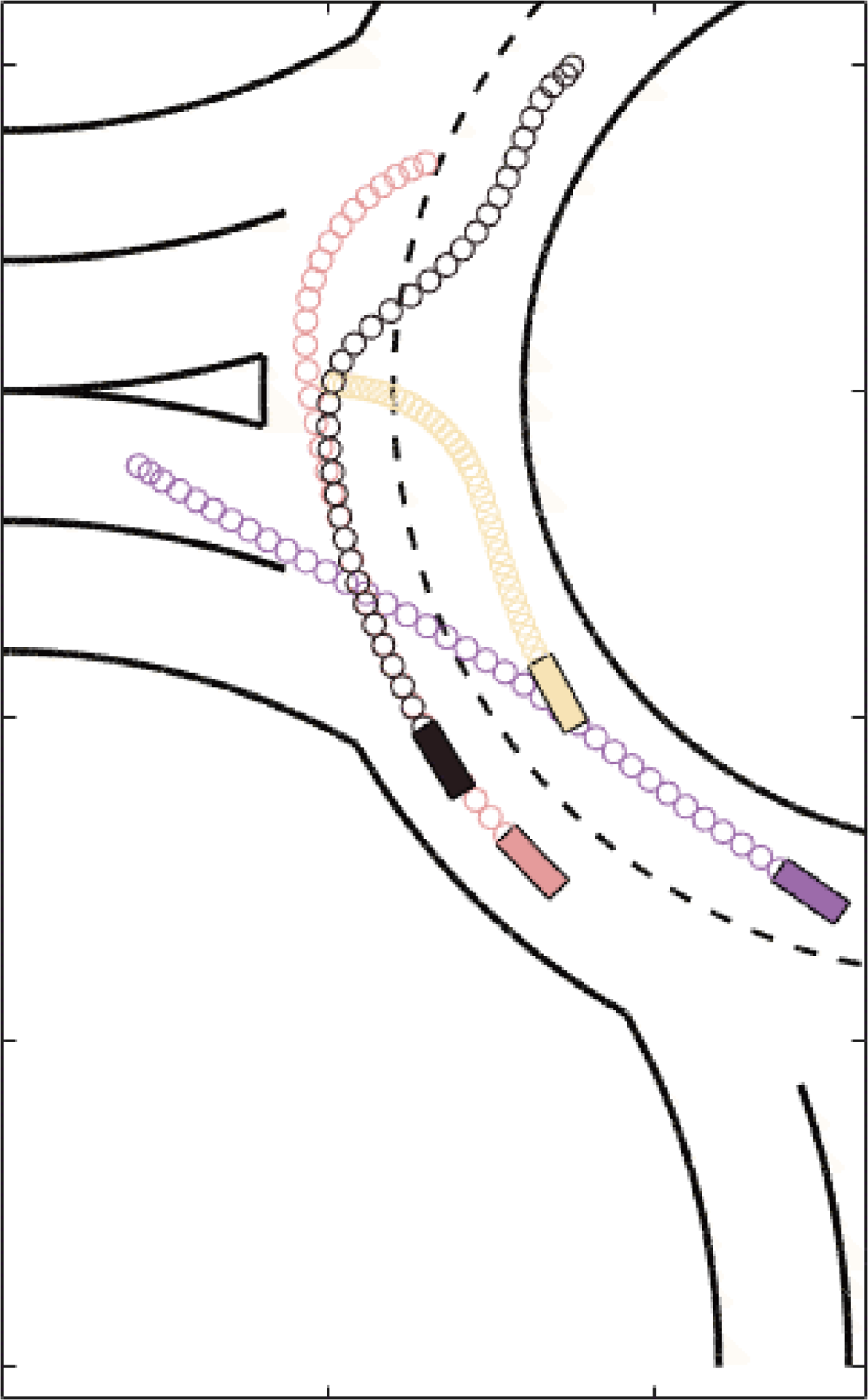}
     \label{Round_3.6}
     } \hspace{1mm}
         \subfloat[$t=5.0$\,s]{\includegraphics[width=0.17\linewidth,height=4.3cm]{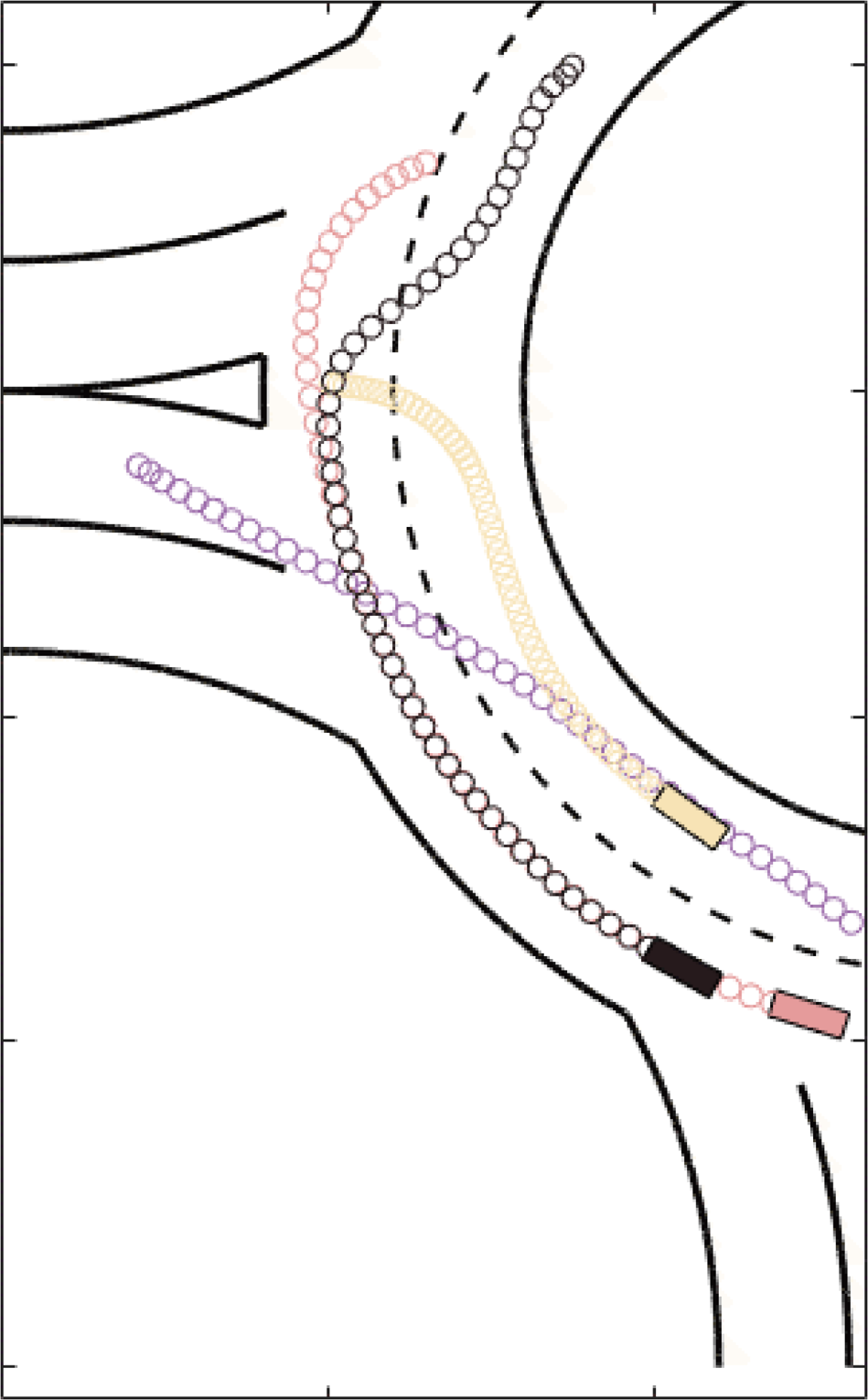}
     \label{Round_5.0}
     }
    \caption{Simulation results for the roundabout scenario.}
    \label{Round}
\end{figure*}
\begin{figure}[t]
    \centering
    \includegraphics[width=0.95\linewidth]{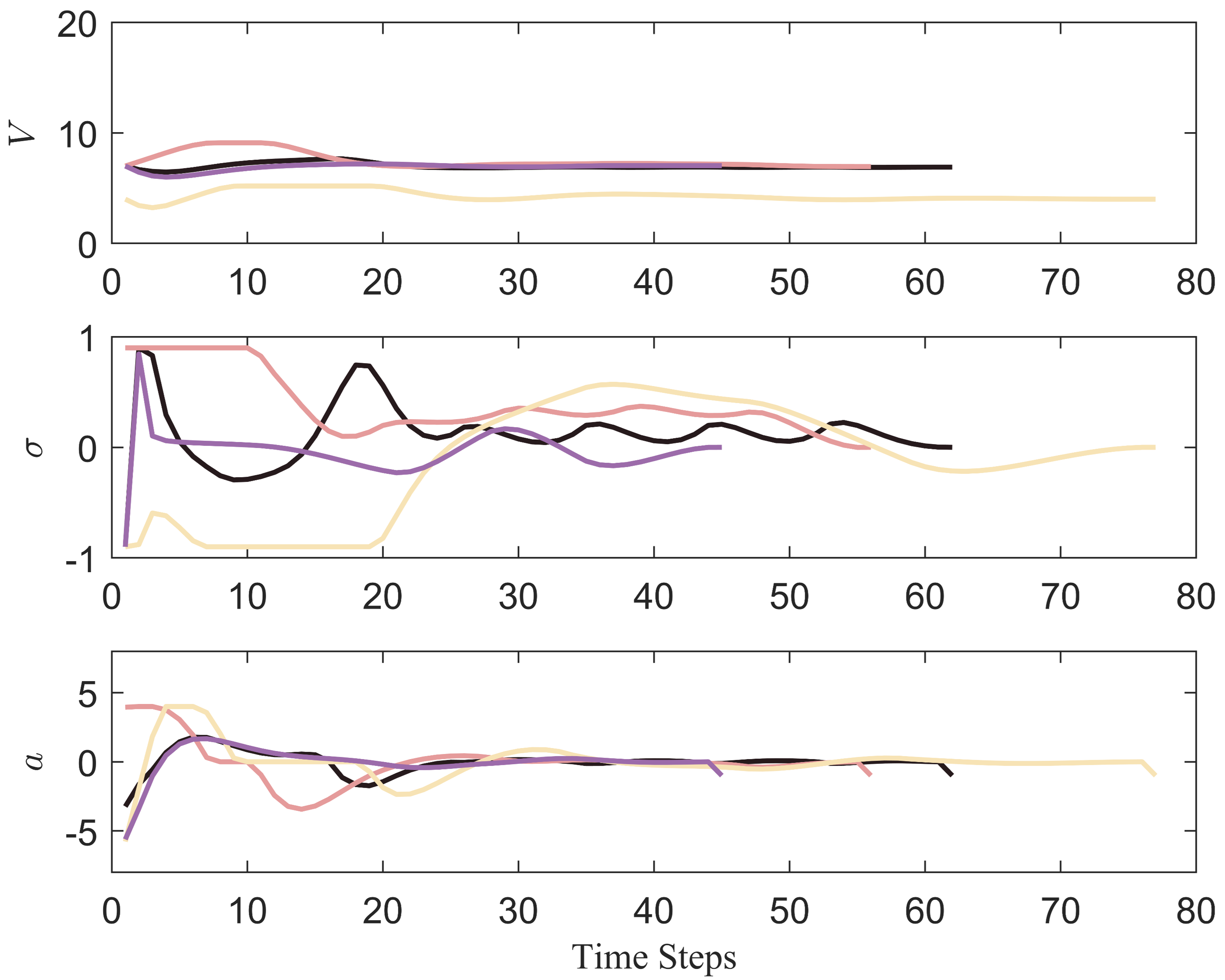}
    \caption{Velocity, acceleration, and steering angle of all CAVs for the roundabout scenario.}
    \label{round-para2}
\end{figure}


Obviously, the optimization order has a impact on the computing time and in some conditions, it influence the effectiveness of the algorithm. In this scenario, by prioritizing optimization of the front CAVs, redundant optimization of strategies can be avoided, resulting in the highest efficiency in problem-solving.

\begin{table}
\centering
\caption{Comparison of computing time and cost value among different methods in different optimization order for the roundabout scenario.
}
\label{Round_time}
\begin{tabular}{|c|c|c|c|}
\hline
Methods & TOPSIS Order & Linear Order & Default Order \\
\hline
Computing Time &$1.56$\,s & $1.93$\,s & $1.95$\,s \\
\hline
$T_{Aver}$ & $0.52$\,s & $0.61$\,s & $0.62$\,s \\
\hline
$J_{Aver}$ & $5.4$ & $5.4$ & $5.4$ \\
\hline
Success Rate   & $25/25$ & $25/25$ & $25/25$ \\
\hline
\end{tabular}
\end{table}

\subsection{Roundabout}
In this section, we demonstrate an example of cooperative decision-making for multiple CAVs on a roundabout. The way-point graph is shown in the Fig. \ref{Round}\subref{Round_Waypoint}, where we sample along the center line of the roads and connect each sampled waypoint with the two nearest waypoints on different lanes in the driving direction to form edges. Fig. \ref{Round}\subref{Round_Initial} shows the initial positions of each CAV. The initial velocities ${V}^{ini}$ of the CAV1-4 are assigned as $7$\,m/s, $7$\,m/s, $4$\,m/s, and $7$\,m/s, with a reference speed set as the initial speed with distribution $\tilde{V}^{ini}$, and the maximum velocity  $V^{max}$ and minimum velocity $V^{min}$ are $0.6\tilde{V}^{ini},1.3\tilde{V}^{ini}$, respectively. 

We randomly assign initial states to all CAVs, and the initial decision strategies are shown in the Fig. \ref{Round}\subref{Round_ini_0.9}-\subref{Round_ini_7}. It can be observed that in the initial decision strategies, there are still unreasonable lane-changing behaviors, such as CAV4 changing lanes downwards at $t=3.5$\,s. The optimized strategy is shown in the Fig. \ref{Round}\subref{Round_1.1}-\subref{Round_5.0} after algorithm processing, the unreasonable lane-changing behaviors are changed, and CAV2 chooses to change lanes while closely following CAV1 with the same speed to ensure that the speed remains stable at the initial speed. At the same time, this strategy reduces the lane-changing behavior of the slower CAV3, so that the velocities of all CAVs require minimal adjustment, as shown in the Fig. \ref{round-para2}, to reduce braking or acceleration actions. The average cost decreased from the initial value of $15.27$ to $8.34$ after $4$ iterations, resulting in a $45\%$ reduction in average cost.

The results of different optimization orders are shown in Table \ref{Round_time}, where default order refers to the sequential arrangement of CAVs from left to right according to their initial positions, as shown in the Fig. \ref{Round}\subref{Round_Initial}. The TOPSIS method still have the best performance. However, the shortcoming of the linear method leads to worse performance.

\begin{table}
\centering
\caption{Comparison of computing time and cost value among different methods in different optimization order for the unsignalized intersection scenario.
}
\label{Inter_time}
\begin{tabular}{|c|c|c|c|}
\hline
Methods & TOPSIS Order & Linear Order & Default Order \\
\hline
Computing Time &$9.21$\,s & $9.23$\,s & $9.22$\,s \\
\hline
$T_{Aver}$ & $3.07$\,s & $3.08$\,s & $3.07$\,s \\
\hline
$J_{Aver}$ & $9.7$ & $9.7$ & $9.7$ \\
\hline
Success Rate   & $25/25$ & $25/25$ & $25/25$ \\
\hline
\end{tabular}
\end{table}

\begin{figure*}[htbp]
    \centering
     \subfloat[Way-point Graph]{\includegraphics[width=0.23\linewidth,height=3.6cm]{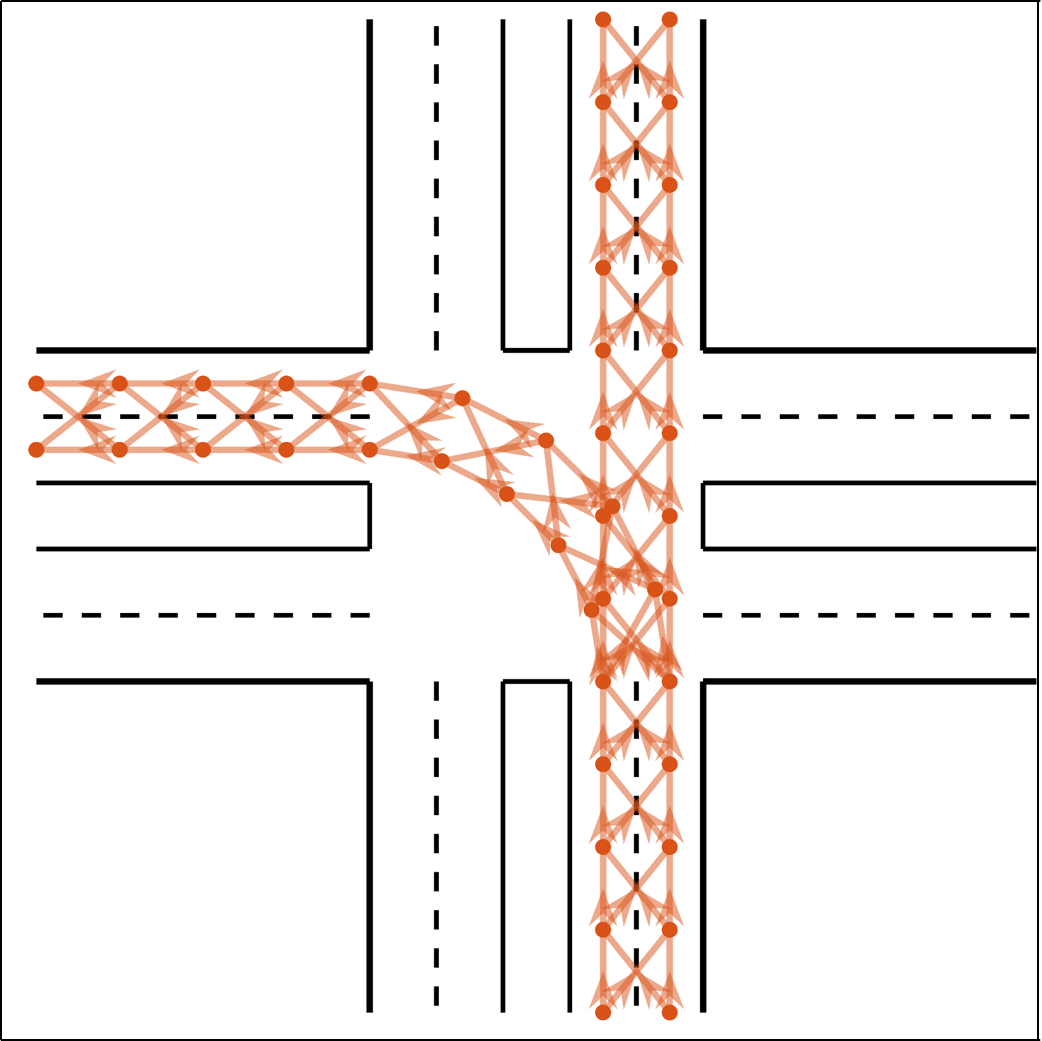}
     \label{Inter_Waypoint}
     }\hspace{1mm}
          \subfloat[$t=1.0$\,s]{\includegraphics[width=0.23\linewidth,height=3.6cm]{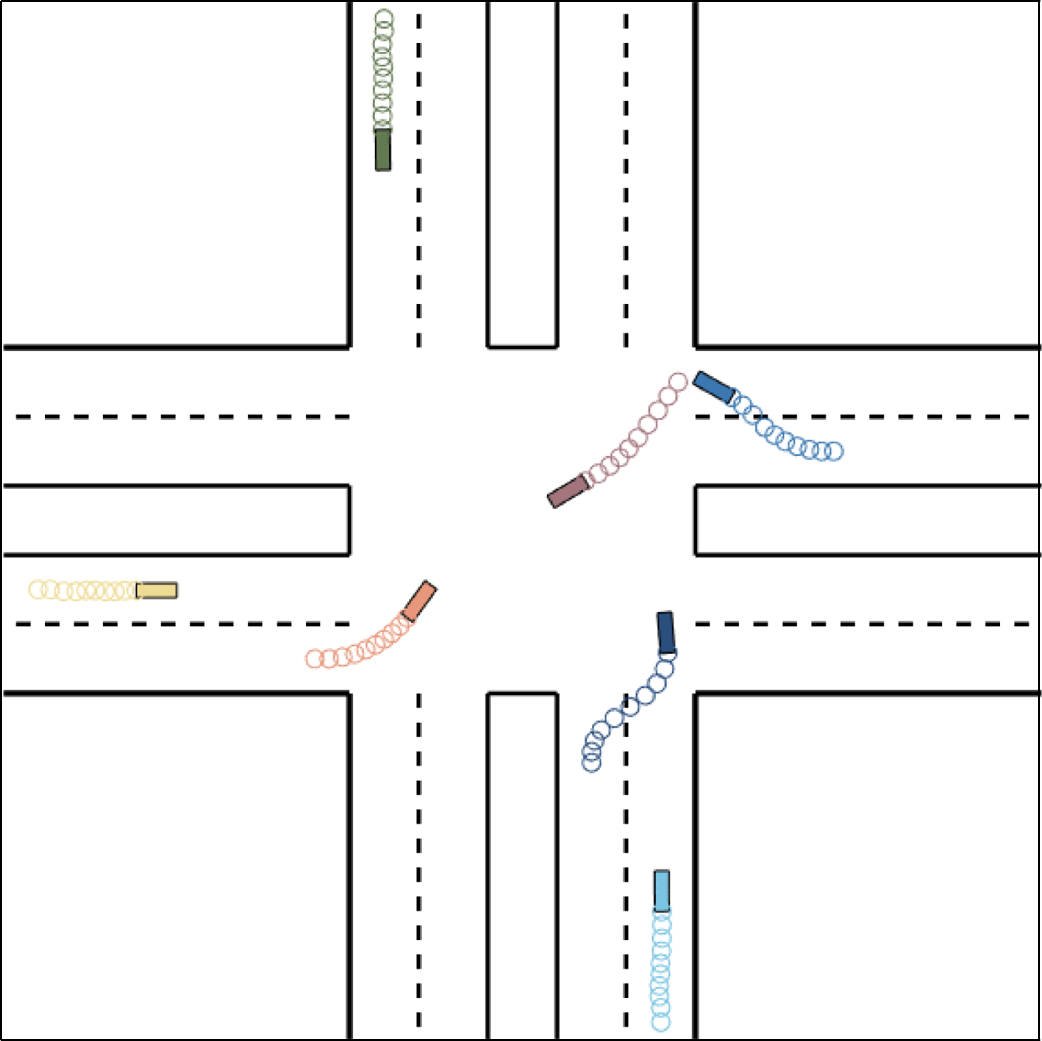}
     \label{Inter_ini_1.0}
     }\hspace{1mm}
          \subfloat[$t=1.6$\,s]{\includegraphics[width=0.23\linewidth,height=3.6cm]{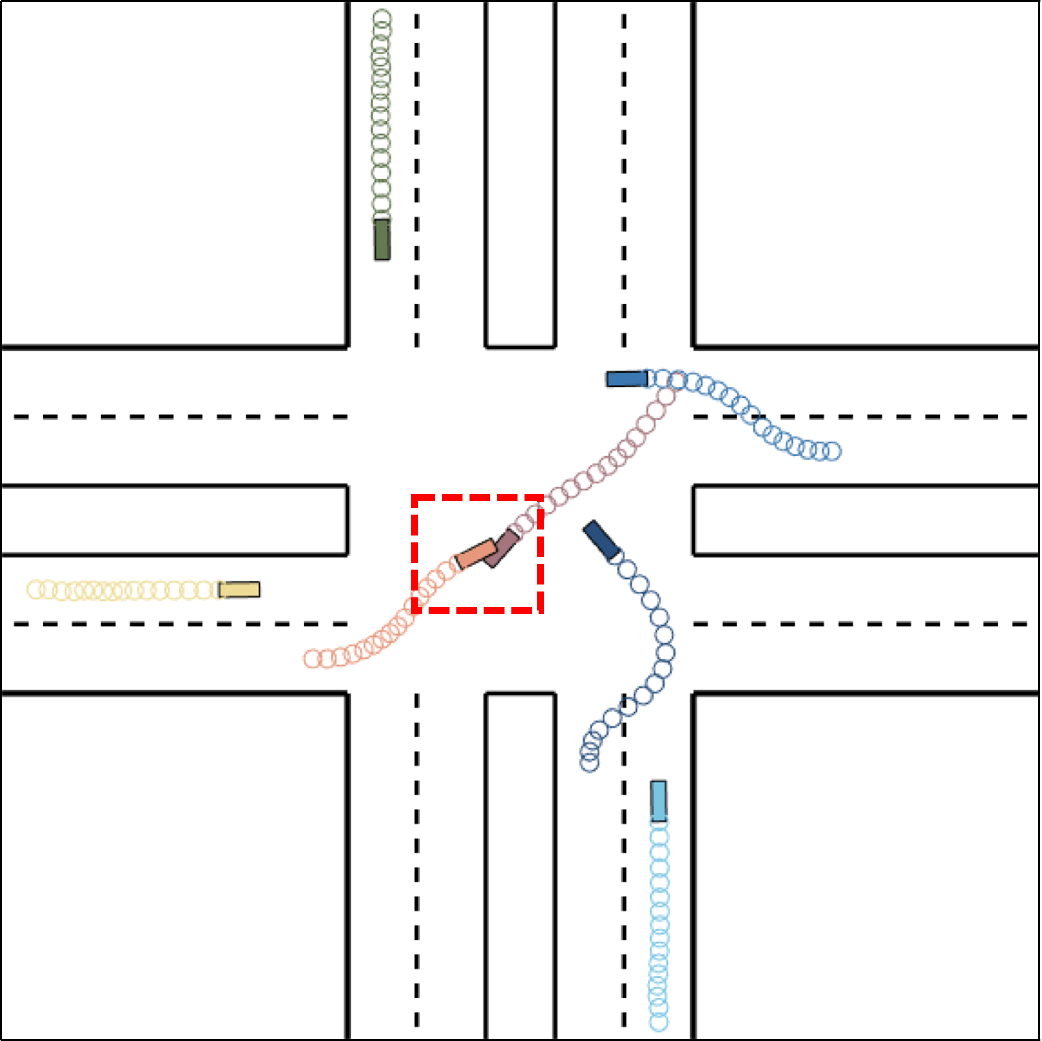}
     \label{Inter_ini_1.6}
     }\hspace{1mm}
          \subfloat[$t=4.1$\,s]{\includegraphics[width=0.23\linewidth,height=3.6cm]{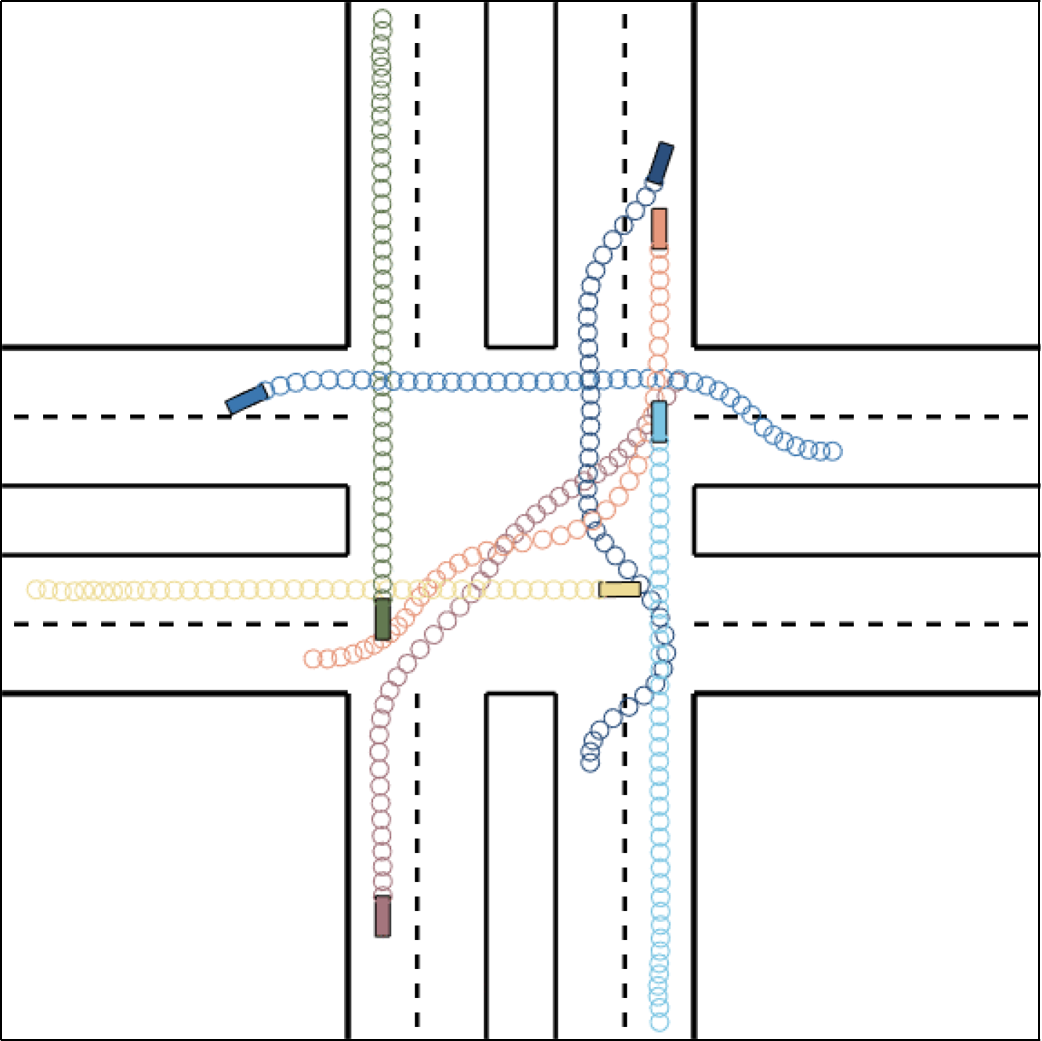}
     \label{Inter_ini_4.1}
     }
     \vspace{-3mm}
     \subfloat[Initial Position]{\includegraphics[width=0.23\linewidth,height=3.6cm]{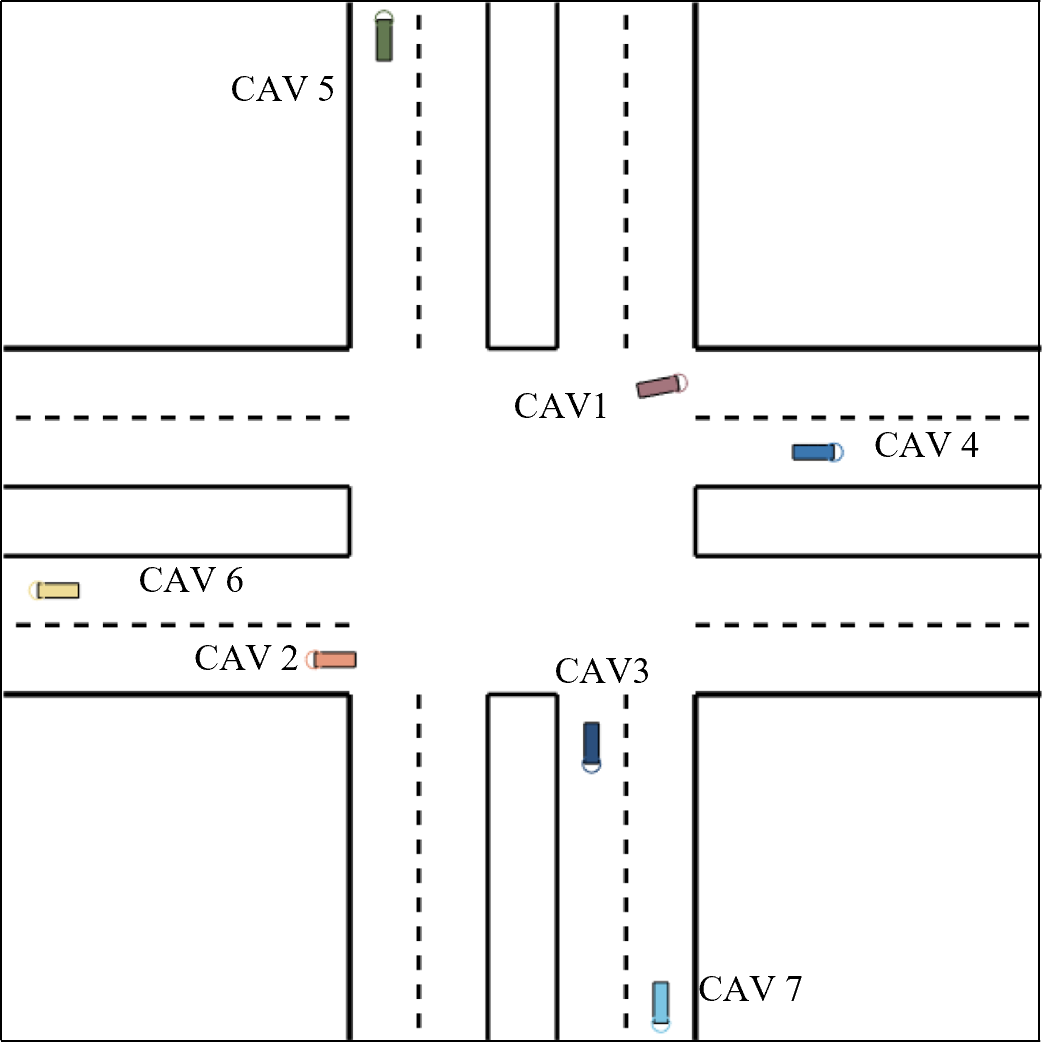}
     \label{Inter_Initial}
     }\hspace{1mm}
               \subfloat[$t=1.6$\,s]{\includegraphics[width=0.23\linewidth,height=3.6cm]{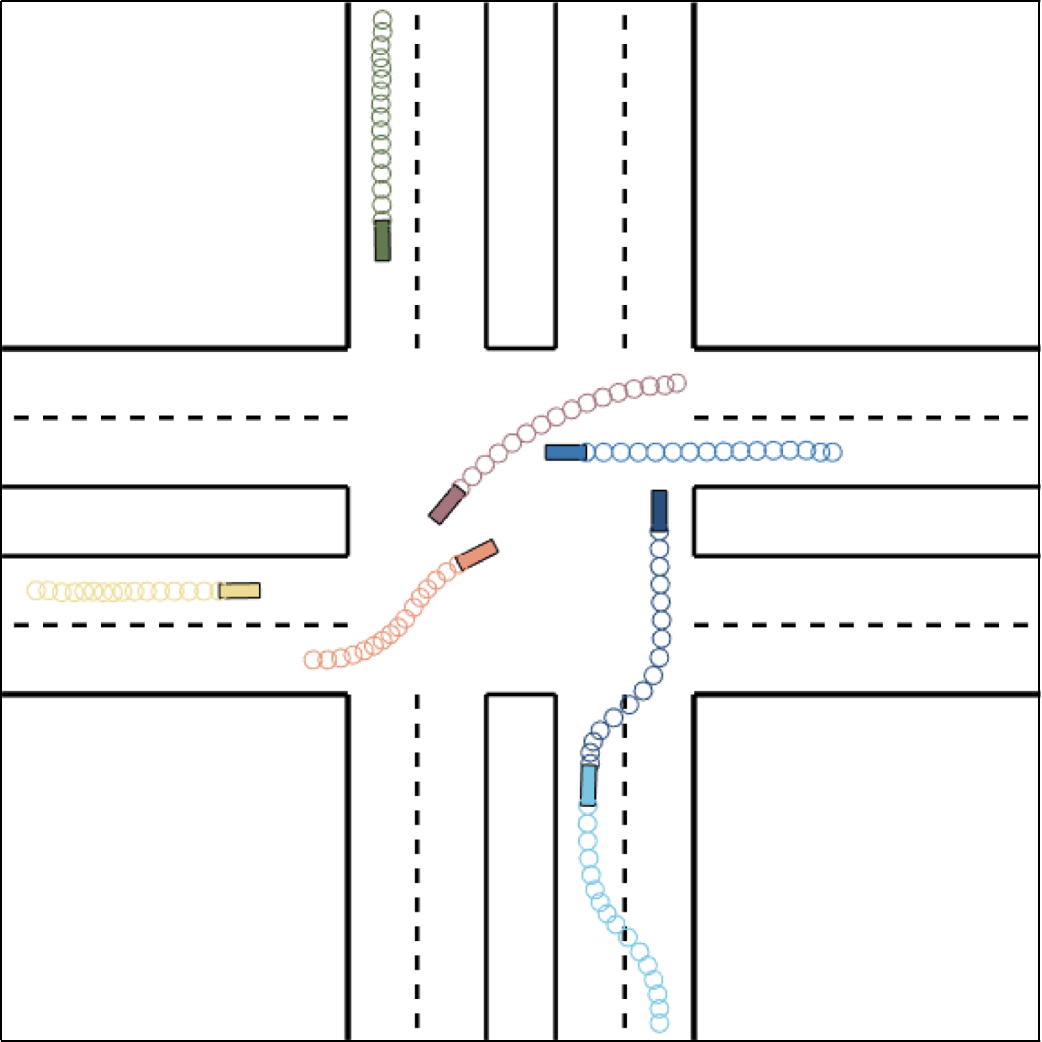}
     \label{Inter_1.6}
     }\hspace{1mm}
               \subfloat[$t=2.6$\,s]{\includegraphics[width=0.23\linewidth,height=3.6cm]{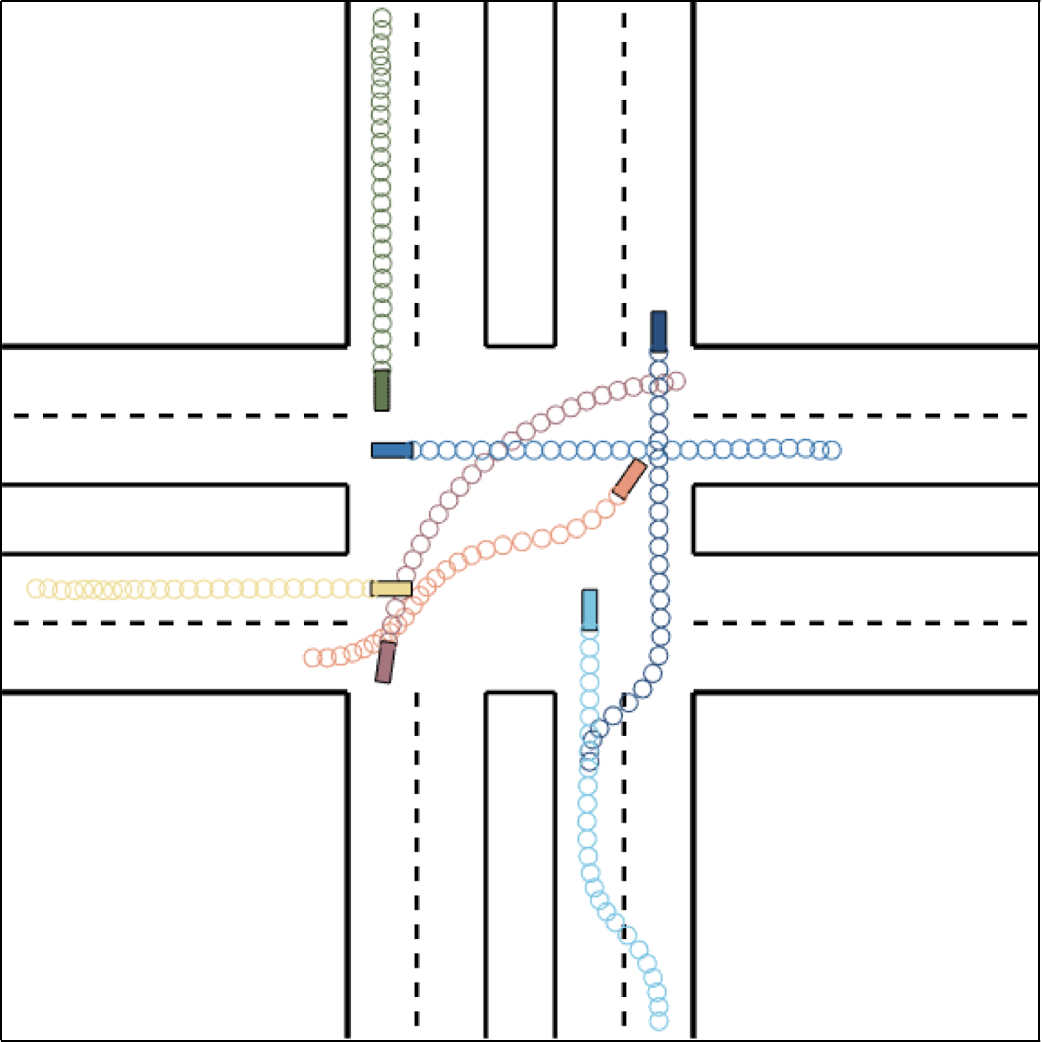}
     \label{Inter_2.6}
     }\hspace{1mm}
               \subfloat[$t=3.9$\,s]{\includegraphics[width=0.23\linewidth,height=3.6cm]{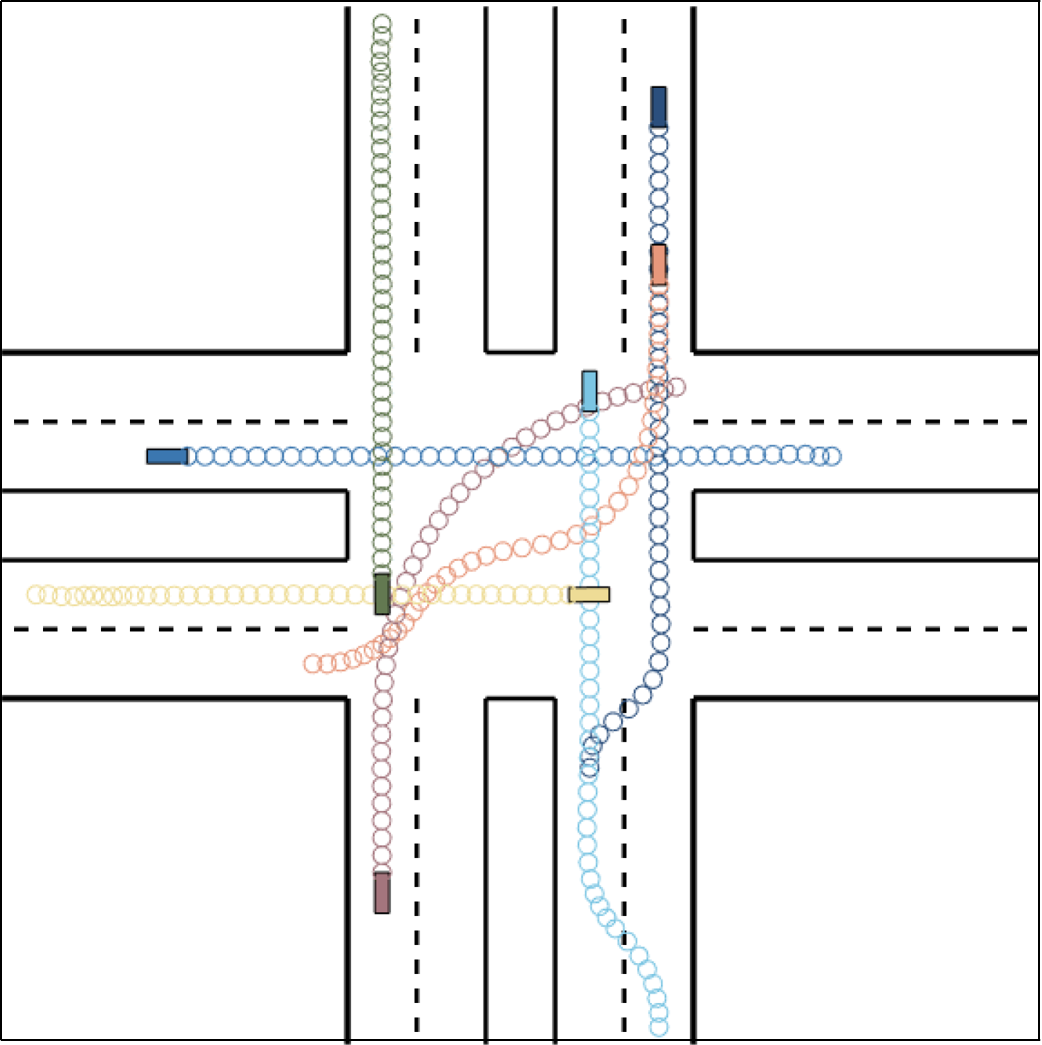}
     \label{Inter_3.9}
     }
     \caption{Simulation results for the unsignalized intersection scenario. 
   }
    \label{inter}
\end{figure*}

\begin{figure}[htbp]
    \centering
    \includegraphics[width=0.95\linewidth]{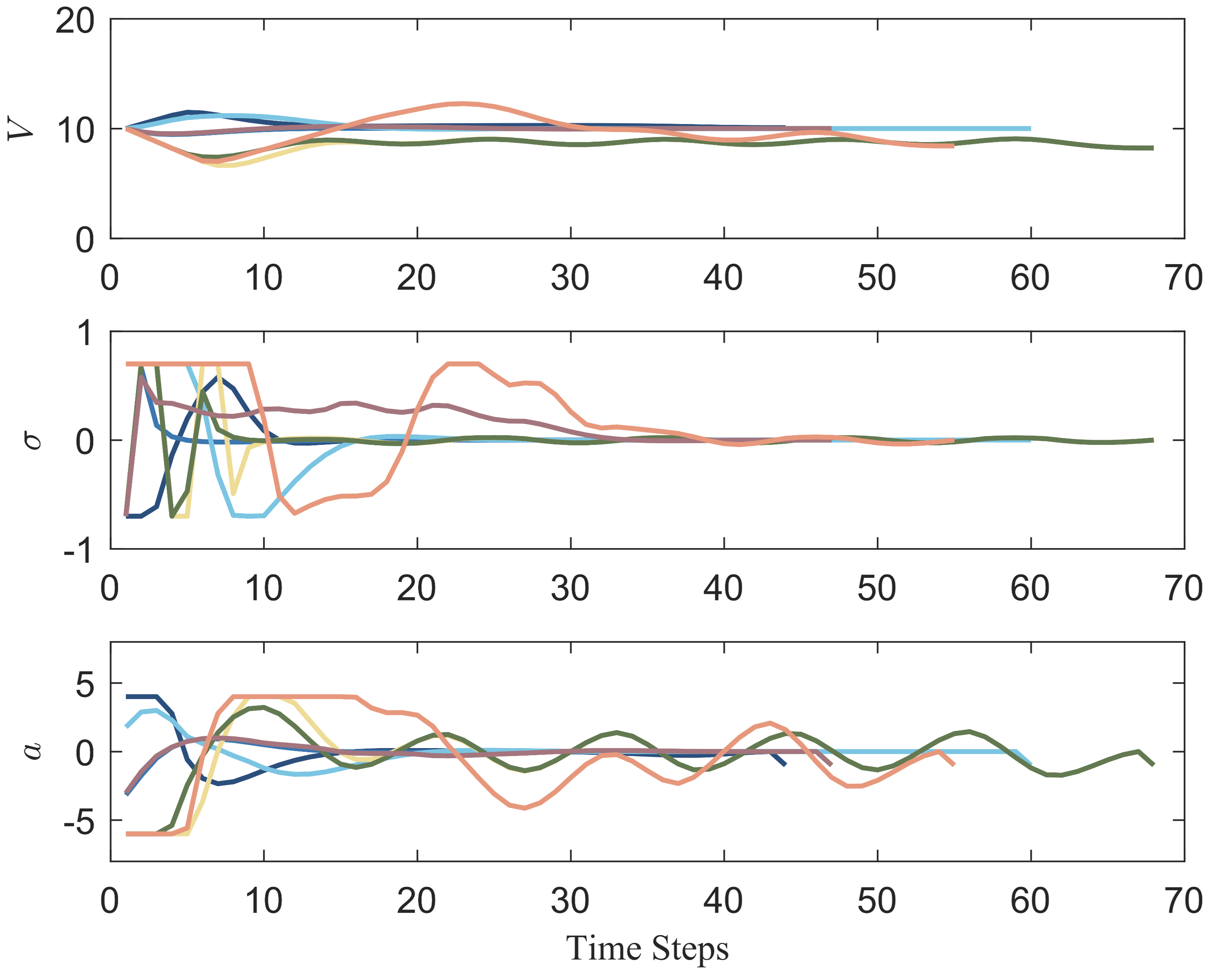}
    \caption{Velocity, acceleration, and steering angle of all CAVs for the unsignalized intersection scenario}
    \label{inter-para2}
\end{figure}
\subsection{Unsignalized Intersection}
In this section, we show an example of cooperative decision-making for multiple CAVs at an unsignalized intersection. In this scenario, we only consider left turn and straight through movements. Each lane is $3.75$\,m wide, and we sample points equidistant along the centerline of the road for the straight segments. Each sampled point is connected to the two nearest points on different lanes in the direction of travel to form edges. To demonstrate the effectiveness of the way-point graph, we only show the straight-through and left-turn movements at the bottom right intersection, as shown in the Fig. \ref{inter}\subref{Inter_Waypoint}. The initial positions of each CAV are shown in the Fig. \ref{inter}\subref{Inter_Initial}. The initial velocities ${V}^{ini}$ of the CAVs are all $10$\,m/s, with a reference speed equal to the initial speed with distribution $\tilde{V}^{ini}$, and the maximum velocity  $V^{max}$ and minimum velocity $V^{min}$ are $0.6\tilde{V}^{ini},1.3\tilde{V}^{ini}$, respectively. 

In this scenario, we assume that CAV6 and CAV7 intend to turn left and other CAVs intend to go stright. We randomly assign an initial state, and the initial strategies are shown in the Fig. \ref{inter}\subref{Inter_ini_1.0}-\subref{Inter_ini_4.1}. In the initial state, the CAV6 and CAV7 are involved in a collision at $t=1.6$\,s and other CAVs exhibit many ineffective and irrational lane-changing behaviors, such as CAV1 at $t=1.6$\,s and $t=4.1$\,s, which can lead to higher fuel consumption and may potentially cause safety accidents. After the algorithm processing, the strategies are adjusted as shown in the Fig. \ref{inter}\subref{Inter_1.6}-\subref{Inter_3.9}, where the CAVs avoid the collision and retain only necessary lane-changing behaviors. For example, at $t=1.6$\,s, CAV6 in the process of turing left and keep the lane while CAV7 chooses to change lanes to avoid collision. From the speed profiles in the Fig. \ref{inter-para2}, it can be observed that CAV5 slows down to avoid CAV2 and CAV4. The average cost decreased from the initial value of $12.73$ to $8.59$ after $4$ iterations, resulting in a $33\%$ reduction in average cost.

The results of the second experiment are shown in Table \ref{Inter_time}. To demonstrate the effectiveness of the method, we correspondingly arrange the positions in the initial sorting, as shown in the Fig. \ref{inter}\subref{Inter_Initial}. Due to the initial velocities being too close and multiple cars being relatively close to their initial positions, the differences in the results obtained by the three methods are not significant.
\subsection{Comparison of Different Decision-Making Models}
In this section, we set up a series of experiments to compare the solution efficiency and quality of the two models, MIPG and MILP. The MIPG problem is solved by the sequential Gauss-Seidel algorithm we proposed before and the MILP problem is solved by the Gurobi. The results are shown in Table \ref{efficiency resluts}. 
\begin{table}
    \centering
    \caption{Comparison of computing time and cost value among different models to obtain decision-making variables for the three different scenarios}
    \label{efficiency resluts}
    \subfloat[Computing Time]{
        \begin{tabular}{|c|c|c|}
        \hline
        Decision-Making Model Type & MILP & MIPG \\
        \hline
        Overtaking & $8.85$\,s & $3.03$\,s \\
        Roundabout & $32.40$\,s & $2.68$\,s \\
        Unsignalized Intersection & $69.13$\,s & $6.83$\,s \\
        \hline
        \end{tabular}
        \label{subtable:computing_time}
    }
    \hspace{1mm}
    \subfloat[Cost Value]{
        \begin{tabular}{|c|c|c|}
        \hline
       Decision-Making Model Type & MILP & MIPG \\
        \hline
        Overtaking & $7.8$ & $11.0$ \\
        Roundabout & $34.7$ & $35.6$ \\
        Unsignalized Intersection & $41.9$ & $44.5$ \\
        \hline
        \end{tabular}
        \label{subtable:cost_value}
    }
\end{table}
It is evident from the experimental results that the Nash equilibrium solution of the MIPG problem and the globally optimal solution of the MILP problem show slight differences in solution quality. The cost value of the two solutions have a slight difference, and as the complexity of the scenario increases, the cost difference decreases. Based on the previous definitions, we can easily find that adjusting the parameter $\epsilon$ can help reduce this gap. However, solving the MILP problem takes much longer time than solving the MIPG problem with the method we proposed, especially as road structures become more complex and the number of CAVs increases.

\section{Conclusion}

This paper proposes a decision-making approach to address the cooperative decision-making problem of CAVs in diverse urban traffic scenarios with a generic road topology. The problem is modeled as an MIPG, and the Nash equilibrium solution is obtained by employing the proposed Gauss-Seidel algorithms for iteration, thereby deriving the trajectories for each CAV. 
To validate the effectiveness of this method, simulations are conducted for three urban traffic scenarios with significantly different topological structures. Simulation results demonstrate that the proposed optimization scheme can efficiently generate more reasonable strategies for CAVs, and the generalization ability of the proposed method is appropriately verified. Furthermore, it is shown that the proposed sequential Gauss-Seidel algorithms can improve the computational efficiency compared to the baseline methods.

\bibliographystyle{ieeetr}
\bibliography{Reference}

\end{document}